\documentclass[10pt,twocolumn,letterpaper]{article}
\usepackage[pagenumbers]{cvpr}

\usepackage{afterpage}
\usepackage{colortbl}

\renewcommand{\paragraph}[1]{\vspace{.5em}\noindent\textbf{#1.}}

\makeatletter
\IfFormatAtLeastTF{2025-06-01}{%
  \@ifpackageloaded{lineno}{%
    \AddToHook{begindocument/end}{%
      \RemoveFromHook{build/column/before}[lineno]%
      \def\@LN@makecol{%
        \@LN@orig@makecol
        \setbox\@outputbox\vbox{%
          \boxmaxdepth\@maxdepth
          \protected@write\@auxout{}{%
            \string\def\string\@LN@column{\if@firstcolumn1\else2\fi}}%
          \box\@outputbox}}%
      \let\@makecol\@LN@makecol}%
  }{}%
}{}
\makeatother

\definecolor{cvprblue}{rgb}{0.21,0.49,0.74}
\usepackage[pagebackref,breaklinks,colorlinks,allcolors=cvprblue]{hyperref}
\newcommand{\piO}{\ensuremath{\pi_0}}
\newcommand{\piH}{\ensuremath{\pi_{0.5}}}

\makeatletter\patchcmd{\@maketitle}{\vskip .375in}{}{}{}\makeatother
\title{Seeing Is Not Reaching: Instruction Shortcuts Override Visual Grounding in Vision-Language-Action Models}
\author{\raisebox{9.4pt}[0pt][21.302pt]{\begin{tabular}[t]{@{}c@{}}
Qing Zhang$^{1}$ \quad Xuesong Li$^{1,2}$ \quad Zhaoyuan Yang$^{3}$ \quad Changlin Song$^{1}$ \quad Xinyu Tian$^{1}$ \quad Jing Zhang$^{1}$\\
$^{1}$Australian National University \quad $^{2}$CSIRO \quad $^{3}$GE Research\\
{\fontsize{7.7}{9}\selectfont\tt \{Qing.Zhang, Xuesong.Li, Changlin.Song, Xinyu.Tian, Jing.Zhang\}@anu.edu.au, zhaoyuan.yang.ge@gmail.com}\\[-2pt]
{\small Code: \url{https://github.com/QZhang2111/psg}}
\end{tabular}}}
\begin{document}
\maketitle
\suppressfloats[t]
\begin{abstract}
Vision-language-action models (VLAs) are expected to inherit visual grounding from vision-language models, yet often reach for an object's training location after it has been moved.
Across six VLAs, we identify two shortcuts: a scene shortcut, where the arm follows the scene's original task despite a new instruction, and an instruction shortcut, where the instruction retrieves the requested object's training location.
In \piH, the instruction shortcut persists even though object-name attention follows the moved target.
We call this \emph{Seeing Is Not Reaching}.
To determine whether this grounding is unused or overridden, we separately mask an image-mediated grounding pathway and a direct instruction-to-action pathway.
Masking the grounding pathway eliminates reaches to moved targets, whereas masking the direct pathway increases them.
Thus, grounding already guides action but is often overridden by the instruction shortcut, which acts through the direct pathway.
We introduce Pathway-Separated Guidance (PSG), which amplifies grounding and suppresses the shortcut without training or external target locations.
On LIBERO-PRO Object under Position perturbations, PSG raises the target-reaching rate from 22.0\% to 52.6\% and task success from 16.6\% to 40.8\%, while largely preserving performance on the original tasks.
\end{abstract}
\begin{figure}[t]
\centering
\includegraphics[width=\linewidth]{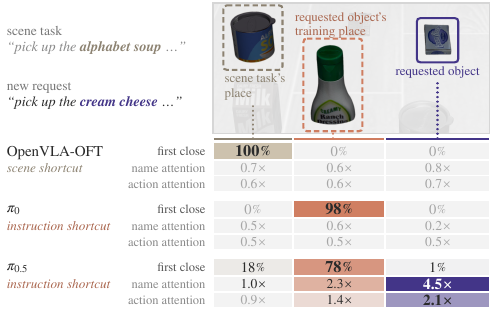}%
\caption{\textbf{\boldmath \piH\ attends to the requested object yet follows an instruction shortcut.}
Under Task-paired (same scene, new request), OpenVLA-OFT follows a scene shortcut, while \piO\ and \piH\ follow an instruction shortcut to the requested object's training place.
First-close rates and image-average-normalized attention are aggregated over five scene--request pairs.}

\label{fig:teaser}
\end{figure}
\section{Introduction}
\label{sec:intro}

Vision-language-action models (VLAs) build on pretrained vision-language models (VLMs)~\cite{zitkovich2023rt2,kim2024openvla,black2024pi0,bjorck2025gr00t,black2025pi05,zheng2025xvla}.
VLMs can locate named objects in images, known as visual grounding~\cite{karamcheti2024prismatic,beyer2024paligemma,kang2025fewheads}.
VLAs are expected to inherit this ability and use it to reach objects at new locations.
Yet benchmarks such as LIBERO-PRO and LIBERO-Plus show sharp drops in success after target relocation, with the arm often returning to the target's training location~\cite{zhou2026liberopro,fei2026liberoplus,li2025vlasconfined}.

Prior studies offer different explanations for these failures.
On the original LIBERO scenes, policies often retain much of their success without instructions~\cite{fei2026liberoplus,fang2026cag}, suggesting reliance on task-correlated visual cues~\cite{hou2026langgap,guo2026robofollow,xing2025shortcut,gerigk2026scrubbing}.
Other studies find that instructions select familiar behaviors, while visual feedback adapts their execution~\cite{li2025vlasconfined,liu2026perturbot,chen2026instructions}.
What drives the arm does not reveal what visual information the model retains.
Robot training can weaken visual and semantic abilities inherited from VLMs~\cite{kachaev2025dontblind,hancock2026vlm2vla,driess2025knowledge,zhang2025howinherit}.
Meanwhile, the target's identity can remain decodable even when the arm reaches for another object~\cite{jiang2026encoded}.
Yet decodable identity does not establish whether the model can locate the target from its name in the image, much less whether that information guides action.
We therefore ask: \emph{When the target moves, is visual grounding lost, retained but unused, or overridden by another effect of the instruction?}

We examine six VLAs under two changes: relocating the target or requesting a different object in the same scene.
Two shortcuts emerge (Fig.~\ref{fig:teaser}): a \emph{scene shortcut}, where the arm goes to the same location despite a changed instruction, and an \emph{instruction shortcut}, where it goes to the requested object's training location despite relocation~\cite{geirhos2020shortcut}.
These failures might reflect weakened visual grounding: in several models, name attention fails to follow the relocated target (Fig.~\ref{fig:retention}).
In \piH, however, name attention follows the moved object, yet the arm still predominantly takes the instruction shortcut.
We call this \emph{Seeing Is Not Reaching}.

To determine whether this apparent grounding is unused or overridden, we separately mask two instruction-to-action pathways (Fig.~\ref{fig:cut}).
The \emph{grounding pathway} passes through image tokens that attend to object names before being read by action tokens~\cite{swann2026causal}.
The \emph{direct pathway} allows action tokens to attend directly to the instruction.
Across 200 relocated-target episodes in a diagnostic test, masking the grounding pathway reduces target reaches from 29 to 0, while masking the direct pathway increases them from 29 to 64.
These opposing effects show that grounding already guides action but is overridden by an instruction shortcut on the direct pathway.
Yet in the original scenes, masking the direct pathway still leaves reaches in 199 of 200 episodes, consistent with prior ablations~\cite{shi2026vlatrace}.
When the target occupies its training location, grounding and the shortcut point to the same place; relocation exposes their conflict.

This conflict suggests a correction: amplify grounding while suppressing the instruction shortcut.
We introduce Pathway-Separated Guidance (PSG), which estimates each pathway's contribution by masking it and measuring the resulting change in predicted flow velocity.
PSG then amplifies the grounding contribution and suppresses the direct-pathway contribution during sampling.
Unlike instruction-free guidance, which mixes both effects~\cite{fang2026cag,zhan2026rss}, PSG adjusts them separately during sampling.
PSG requires neither training nor externally supplied target locations.
On LIBERO-PRO Object under Position perturbations, PSG raises \piH's target-reaching rate (Reach) from 22.0\% to 52.6\% and task success from 16.6\% to 40.8\%, while largely preserving original-task performance.

\noindent{In summary, our contributions are as follows:}
\begin{itemize}[leftmargin=*,itemsep=0pt,topsep=2pt]
\item Target relocation and instruction changes reveal two distinct shortcuts across six VLAs. In \piH, name attention follows the moved target while the arm still takes the instruction shortcut, which we call \emph{Seeing Is Not Reaching}.
\item Pathway masking shows that grounding already guides action in \piH\ but is overridden by the instruction shortcut on the direct pathway. Nearly dispensable in familiar scenes, this pathway contributes to errors after relocation.
\item Pathway-Separated Guidance (PSG) adjusts these effects separately without training or external target locations. It more than doubles both Reach and task success on LIBERO-PRO Object under Position perturbations.
\end{itemize}

\afterpage{%
\begin{figure*}[t]
\centering
\includegraphics[width=\linewidth]{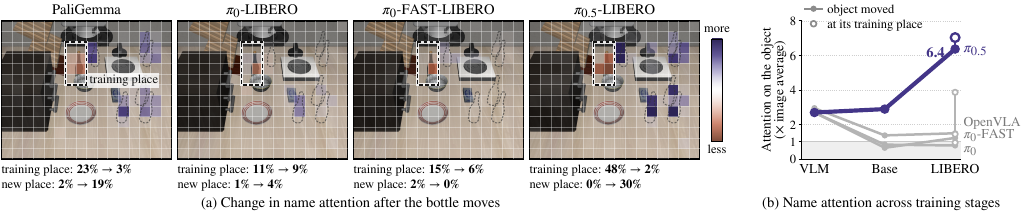}%
\caption{\textbf{\boldmath Name attention follows the moved object along \piH's lineage.}
(a)~Change in name-token attention over main-camera patches after relocating a wine bottle. PaliGemma is the shared VLM backbone before robot training; among the three $\pi$ policies, only \piH\ shows a similar shift toward the moved object.
(b)~Name attention to the moved object, relative to the image average, across VLM, robot-pretrained base, and LIBERO-fine-tuned stages, averaged over nine objects. OpenVLA is included for comparison.}
\label{fig:retention}\label{fig:lineage}
\end{figure*}%
}

\section{Related Work}

\label{sec:related}

\noindent\textbf{Shortcuts in VLAs.}
Robot demonstrations often couple scenes, instructions, and actions, allowing VLAs to exploit task-correlated shortcuts~\cite{dehaan2019causal,geirhos2020shortcut,xing2025shortcut}.
Counterfactual benchmarks expose these dependencies by relocating objects, changing instructions, or perturbing scene cues~\cite{zhou2026liberopro,fei2026liberoplus,zhou2026liberopro2,gerigk2026scrubbing}.
One pattern is scene dependence: policies can execute familiar tasks even when instructions are absent or changed~\cite{fei2026liberoplus,fang2026cag,guo2026robofollow,jiang2026encoded,jiang2026benchmarking}.
Another is instruction-conditioned retrieval: instructions select familiar trajectories or locations associated with demonstrated tasks, while vision adapts their execution~\cite{li2025vlasconfined,chen2026instructions,liu2026perturbot}.
Replacing the instruction-conditioned prefix of \piH\ with that of another task redirects the arm to that task's object~\cite{chen2026instructions}.
These dependencies need not be mutually exclusive, and similar failures can arise from different cues.
Identifying what selects the behavior, however, does not establish whether the model can still locate the named object in the current image.

\noindent\textbf{How language reaches the action.}
Robot training can weaken visual and semantic abilities inherited from VLMs~\cite{kachaev2025dontblind,hancock2026vlm2vla,driess2025knowledge,zhang2025howinherit,kachaev2026act2answer}.
Yet target identity and executed-task information can remain decodable from fine-tuned representations~\cite{jiang2026encoded,chen2026instructions}.
Jiang et al.~\cite{jiang2026encoded} describe this as a gap between what is encoded and what controls behavior, without identifying its cause.
Decodability alone does not establish whether the model locates the named object in the image or uses it to guide action.
To examine causal use, other studies intervene on how information enters action generation~\cite{haon2025steeringvla,grant2026features}.
Swann et al.~\cite{swann2026causal} show that image tokens in \piH\ can carry instruction-conditioned target locations.
Patching these tokens from a run with another instruction can redirect reaching to the other target.
The same study also finds target-location information in later text and readout tokens, so it is not confined to image tokens.
VLA-Trace~\cite{shi2026vlatrace} separately masks prefix vision--language exchange and the action expert's direct access to instructions.
On original LIBERO tasks, removing direct access has little effect on success, whereas disrupting prefix exchange can be far more damaging.
These studies establish causal routes for instruction information, but leave open how each pathway affects where the arm reaches after the target is relocated.

\noindent\textbf{Correcting shortcuts.}
Counterfactual demonstrations and representation constraints can reduce shortcut reliance during training~\cite{glossop2025cast,chen2026instructions,gerigk2026scrubbing,xu2026apt,liu2026perturbot,kondoh2026counteralign}.
Other methods steer frozen policies using directions from labeled rollouts or externally supplied target information~\cite{lei2026roboirs,zhang2026reguide,song2026omniguide,liu2026vls}.
Inference-time guidance modifies predictions by contrasting a conditioned output with a reference~\cite{ho2022cfg,karras2024autoguidance,park2026acg,wu2026pcd}.
For VLAs, CAG and RSS strengthen language conditioning relative to vision-only or instruction-free predictions~\cite{fang2026cag,zhan2026rss}, while IGAR reallocates attention toward instruction tokens~\cite{zhang2026igar}.
However, a whole-instruction contrast changes both the image-mediated grounding pathway and the direct instruction pathway, leaving their effects mixed in one guidance signal.
When these effects conflict, stronger language conditioning need not improve reaching of the current target.
PSG instead builds separate pathway-specific velocity contrasts, amplifying grounding while suppressing the direct instruction shortcut during sampling.
It needs neither additional training nor external target locations, and we compare it with CAG on the same model in \cref{sec:experiments}.
\section{Seeing Is Not Reaching}
\label{sec:mechanism}

When a VLA reaches for an object's training place rather than its current position, the behavior alone does not reveal what determines the arm's destination.
We first distinguish the roles of the scene and instruction, then examine whether visual grounding is retained and influences action.

\subsection{Scene Shortcut or Instruction Shortcut?}
\label{sec:behavior}
We compare six VLAs on five LIBERO-Object tasks~\cite{liu2023libero}.
Instead of task success, we record where the gripper first closes, and call a close on the requested object a \emph{reach}.

\noindent\textbf{Five VLAs return to the training place.}
The \emph{Position} perturbation of LIBERO-PRO~\cite{zhou2026liberopro} swaps the requested object with another object while keeping the instruction unchanged.
Five models then close almost exclusively at the requested object's training place, now occupied by another object  (\cref{tab:pro}; see supplementary).
\piH\ also usually returns there, but reaches the moved object in 42 of 200 episodes.

\afterpage{%
\begin{figure*}[t]
\centering
\includegraphics[width=\linewidth]{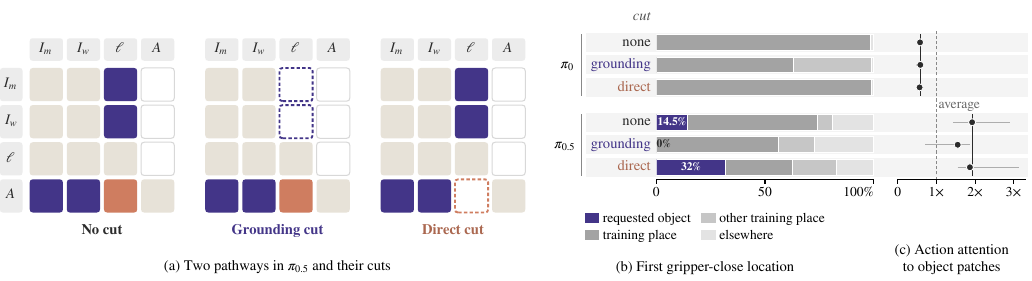}%
\caption{\textbf{\boldmath Pathway cuts have opposite effects on \piH's reaching.}
(a)~Two instruction-to-action pathways: through image tokens (purple) or directly to action tokens (orange). $I_m$, $I_w$: main- and wrist-camera image tokens; $\ell$: instruction; $A$: action. Rows attend to columns; dashed blocks indicate masked connections.
(b)~First gripper-close locations under Position for \piO\ and \piH, with no cut, grounding cut, or direct cut.
(c)~Action-token attention to the moved object's patches, at the first query, relative to the image average (dashed $1\times$ line).}
\label{fig:masks}\label{fig:cut}
\end{figure*}%
}

This result alone cannot tell whether the return follows the scene or the instruction: under Position, the object's training place coincides with the scene's original task location.
We therefore use the Task perturbation of LIBERO-PRO, which keeps the scene unchanged but requests another object in it.
A scene-dependent policy returns to the original task's location, whereas a policy that retrieves a location from the instruction goes to the requested object's training place.
A visually grounded policy instead reaches the requested object itself, at its current location in the image.

\noindent\textbf{Three VLAs follow the scene.}
Under Task-paired, OpenVLA-OFT~\cite{kim2025oft}, TurboVLA~\cite{xie2026turbovla}, and SmolVLA~\cite{shukor2025smolvla} return to the place of the scene's original task despite the changed instruction (\cref{fig:teaser}; see supplementary).
This matches the \emph{scene shortcut} reported in prior work, where policies execute the task familiar from the scene even when the instruction is changed or absent~\cite{xing2025shortcut,fei2026liberoplus,fang2026cag,guo2026robofollow,jiang2026encoded}.
Here, the instruction barely changes where the arm goes.

\noindent\textbf{\boldmath The $\pi$ family follows the instruction, not the object.}
By contrast, \piO, \piO-FAST, and, in most episodes, \piH\ move to the training place of the newly requested object (\cref{fig:teaser}).
For these models, the instruction does change where the arm goes, but it selects a familiar place rather than the object in the current image.
We call this the \emph{instruction shortcut}, in line with reports that instructions retrieve trajectories or locations from demonstrations~\cite{li2025vlasconfined,chen2026instructions,liu2026perturbot}.

Yet this shortcut alone cannot explain why \piH, unlike \piO, sometimes reaches the moved object under Position.
This suggests that \piH\ may retain the ability to locate the named object in the image, which we examine next.

\subsection{Grounding Lost or Grounding Retained?}
\label{sec:retention}
Robot training can weaken visual grounding inherited from VLMs~\cite{kachaev2025dontblind,hancock2026vlm2vla,zhang2025howinherit}, raising the question of whether \piH\ still locates the named object after relocation.
We examine attention from object-name tokens to main-camera image patches before and after moving one object.
In the original scene, the object and its training place coincide; relocation separates them.
We can then test whether name attention follows the object or remains at its training place.

\noindent\textbf{\boldmath Only \piH\ retains the VLM-like attention shift.}
We take PaliGemma~\cite{beyer2024paligemma}, the shared VLM backbone of the three $\pi$ models before any robot training, as a reference.
For example, when the wine bottle is relocated, PaliGemma's name-token attention shifts from the training place to the bottle's new position (\cref{fig:retention}a).
Among the three robot-trained models, only \piH\ shows a similar shift, with attention on the relocated object rising from 0\% to 30\%.
In \piO\ and \piO-FAST, name attention does not follow the bottle to its new position.
Does this divergence emerge during robot pretraining, or only after LIBERO fine-tuning?

\noindent\textbf{The divergence predates LIBERO fine-tuning.}
To trace this divergence, we compare the original VLM, robot-pretrained base models, and LIBERO-fine-tuned policies across nine objects from two suites (\cref{fig:lineage}b).
In PaliGemma, name attention to the moved object is well above the image average.
After robot pre-training, the \piH\ base model~\cite{black2025pi05} still attends to the moved object at a similar level.
By contrast, the \piO\ and \piO-FAST base models~\cite{black2024pi0,pertsch2025fast} have lost this pattern, with name attention falling below the image average.
All three are then fine-tuned on the same LIBERO demonstrations.
After fine-tuning, \piH's name attention reaches 6.4 times the image average, while both \piO\ models remain near the average.
The difference thus already exists before LIBERO fine-tuning begins.

In \piH, name attention follows the relocated object, yet the arm still  mostly returns to the object's training place.
We call this phenomenon \emph{Seeing Is Not Reaching}.
Does this location information remain unused, or does it already influence action without determining where the arm goes?

\subsection{Grounding Unused or Grounding Overridden?}
\label{sec:cut}
To distinguish these possibilities, we separately intervene on two pathways from instruction to action (\cref{fig:masks}a).
In \piH's bidirectional prefix, image tokens attend to instruction tokens and are subsequently read by action tokens~\cite{beyer2024paligemma,swann2026causal}.
We call this the \emph{grounding pathway}.
Action tokens can also attend directly to instruction tokens, forming a \emph{direct pathway} that bypasses the image.
We mask each pathway separately: the grounding cut blocks image tokens from reading the instruction, while the direct cut blocks action tokens from doing so (and also the state token in \piO).
The image and the instruction remain unchanged under both cuts, and each cut leaves the other pathway intact.

The two possibilities make different predictions under pathway cuts.
If grounding is unused, cutting its pathway should leave reaches unchanged; if it guides action, cutting it should reduce reaches.
If the direct pathway counteracts grounding, cutting that pathway should instead increase reaches.
We test these predictions under Position, using a conservative reach criterion.
In two of the five tasks, the relocated object occupies another task's training place, so a close there could reflect a shortcut rather than grounding.
We exclude such closes, so the unmodified \piH\ reaches the moved object in 29 of 200 episodes, compared with the 42 of 200 counted under the looser criterion in \cref{sec:behavior}.

\noindent\textbf{Grounding already guides action.}
Cutting the grounding pathway reduces \piH's reaches from 29 to 0 (\cref{fig:cut}b).
The same cut also reduces the action tokens' attention to the relocated object's image patches (\cref{fig:cut}c).
Thus, the grounding pathway affects both the action tokens' attention to the moved object and where the gripper closes.
With the name-attention shift in \cref{sec:retention}, these results indicate that retained visual grounding already contributes to reaching.

\noindent\textbf{The direct pathway pulls toward the training place.}
Cutting the direct pathway increases \piH's reaches from 29 to 64, while nearly halving closes at the training place (\cref{fig:cut}b).
The action tokens' attention to the moved object's patches changes little despite this increase (\cref{fig:cut}c).
Together, the two cuts reveal opposing effects of the same instruction.
The grounding pathway supports reaching the moved object, while the direct pathway favors its training place.
Thus, grounding is used but overridden by the instruction shortcut carried through the direct pathway.
In \piO, by contrast, neither cut produces reaches to the moved object, consistent with its weakened name attention (\cref{sec:retention}).

\noindent\textbf{Familiar scenes conceal the shortcut.}
In the original scenes, \piH\ with the direct pathway cut still reaches the object in 199 of 200 episodes, consistent with prior ablations~\cite{shi2026vlatrace}.
This contrasts with its strong effect after relocation.
When the object remains at its training place, grounding and the shortcut favor the same destination.
Relocation breaks this agreement, exposing a pathway that pulls the arm from the moved object back to its training place.

Cutting the direct pathway alone, however, yields reaches in only 64 of 200 episodes.
Removing the shortcut is therefore not enough.
The opposing pathway effects suggest a further correction: amplify grounding while suppressing the shortcut during action generation.
This motivates Pathway-Separated Guidance (\cref{sec:framework}).
\section{The Proposed Method}
\label{sec:framework}
The pathway cuts in \cref{sec:cut} reveal opposing effects on reaching.
We turn these diagnostic cuts into guidance signals by measuring how each changes the predicted flow velocity during action generation.
We introduce Pathway-Separated Guidance (PSG), which amplifies the grounding effect and suppresses the direct effect during sampling, without additional training or external target locations.

\subsection{Pathway Effects}
\label{sec:preliminaries}

\piH\ generates action chunks~\cite{zhao2023act,chi2023diffusion} through flow matching~\cite{lipman2023flow,black2024pi0}.
At each replan, the action expert maps Gaussian noise $\mathbf{A}_t^0$ to a chunk $\mathbf{A}_t^1$ through Euler steps:
\begin{equation}
\mathbf{A}_t^{\tau+\delta}
=\mathbf{A}_t^\tau+\delta\,v,
\qquad
v=v_\theta(\mathbf{A}_t^\tau,\mathbf{o}_t),
\label{eq:generation}
\end{equation}
where $\tau\in[0,1]$ is the flow timestep, $\delta$ is the step size, and $\mathbf{o}_t$ contains the main- and wrist-camera images and the instruction.
The initial noise $\mathbf{A}_t^0$ is independent of the instruction, and therefore the instruction affects the generated action chunk only through the predicted velocity $v$.

To measure a pathway's effect on $v$, we compare predictions with and without that pathway at the same sampling state.
Let $I_m$ denote the main-camera image tokens, $\ell$ the instruction tokens, and $A$ the action tokens.
We use $N\subset\ell$ for the \emph{name tokens}, those of every noun phrase in the instruction, identified by a fixed text rule.
We write $v[X\nrightarrow Y]$ for the velocity predicted when tokens $X$ cannot attend to tokens $Y$.
Each cut applies across all layers and attention heads, while the observation, noisy action, and flow timestep remain unchanged.
We define the \emph{grounding effect} and the \emph{direct effect} as
\begin{equation}
v_g=v-v[I_m\nrightarrow N],
\qquad
v_d=v-v[A\nrightarrow\ell].
\label{eq:velocities}
\end{equation}

The direct cut $[A\nrightarrow\ell]$ is the same as in \cref{sec:cut}.
For grounding, rather than blocking both cameras from the entire instruction, we block only main-camera image tokens from reading name tokens.
This choice follows \cref{sec:retention}, where name attention to the moved object was measured in the main-camera image.
All other attention remains unmasked; \cref{sec:where} examines the role of the wrist camera.

\begin{figure}[t]
\centering
\includegraphics[width=\linewidth]{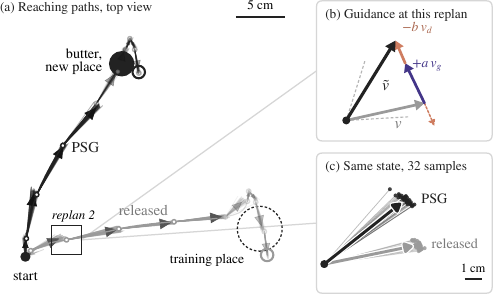}%
\caption{\textbf{PSG redirects reaching toward the moved object.}
(a)~Top-view reaching paths of the released policy (gray) and PSG (black). The released policy returns to the training place, while PSG reaches the relocated butter.
(b)~At the highlighted replan, PSG adjusts the predicted flow velocity from $v$ to $\tilde v=v+a v_g-b v_d$.
(c)~Action predictions from 32 samples at the same state, without and with PSG.}
\label{fig:method}
\end{figure}

\subsection{Pathway-Separated Guidance}
\label{sec:shortcut-removal}

Inference-time guidance modifies predictions by contrasting the model's output with a reference.
Existing methods such as CAG and RSS use instruction-free or vision-only predictions as references~\cite{fang2026cag,zhan2026rss}.
In velocity space, instruction-free guidance takes the form
\begin{equation}
\tilde v=v+w\bigl(v-v[\varnothing]\bigr),
\label{eq:uniform}
\end{equation}
where $v[\varnothing]$ is predicted without the instruction.
Removing the instruction affects both the grounding and direct pathways, so the contrast mixes their effects.
Since the two pathways have opposing effects on reaching after the object is relocated (\cref{sec:cut}), a single guidance coefficient $w$ cannot amplify one effect while suppressing the other.

PSG instead adjusts the two pathway effects defined in \cref{eq:velocities} separately.
At every Euler step, it replaces the predicted velocity with
\begin{equation}
\tilde v=v+a\,v_g-b\,v_d,\qquad a,b\ge 0,
\label{eq:guidance}
\end{equation}
where $a$ amplifies the grounding effect and $b$ suppresses the direct effect; $a=b=0$ recovers the original sampler.

The direct-pathway cut improved reaching after relocation while preserving nearly all reaches in familiar scenes (\cref{sec:cut}).
We therefore set $b=1$, subtracting the full direct effect.
Substituting the definition of $v_d$ gives the PSG sampling step:
\begin{equation}
\mathbf{A}_t^{\tau+\delta}
=\mathbf{A}_t^\tau+\delta\,\tilde v,
\qquad
\tilde v=v[A\nrightarrow\ell]+a\,v_g.
\label{eq:base}
\end{equation}
This is \cref{eq:generation} with $v$ replaced by $\tilde v$.
The first term of $\tilde v$ is the velocity predicted with the direct pathway cut.
The second amplifies the grounding effect measured from the original model (\cref{fig:method}).
PSG thus amplifies a grounding effect derived from the model itself, rather than a target direction supplied externally.
\Cref{sec:where} examines when this effect is sufficient for the arm to reach the moved object.

At each Euler step, PSG evaluates $v$, $v[I_m\nrightarrow N]$, and $v[A\nrightarrow\ell]$ at the same noisy action and flow timestep.
Guidance modifies only the six motion dimensions, leaving the gripper velocity unchanged.
Compared with the released policy, PSG requires one additional prefix computation per action chunk and two additional action-expert evaluations per Euler step.
It updates no parameters and uses no object coordinates, segmentation masks, or detector.
\providecommand{\trainedrow}{\color{black!55}\hypersetup{allcolors=black!55}}
\providecommand{\tbd}{\textcolor{gray!70}{--\,--}}
\providecommand{\nb}[1]{\makebox[2.25em][r]{#1}}
\providecommand{\ml}[1]{\makebox[2.25em][c]{\scriptsize\itshape #1}}
\begin{table*}[t]
\footnotesize
\centering
\setlength\heavyrulewidth{0.25ex}
\renewcommand{\arraystretch}{1.0}
\setbox0\hbox{\piH\ + CAG~\cite{fang2026cag}}\dimen0=\wd0
\setbox0\hbox{OpenVLA-OFT~\cite{kim2025oft}}\ifdim\wd0>\dimen0 \dimen0=\wd0 \fi
\setbox0\hbox{\piH\ + PSG (Ours)}\ifdim\wd0>\dimen0 \dimen0=\wd0 \fi
\setbox0\hbox{LIBERO~\cite{liu2023libero}}\advance\dimen0 by \wd0
\setbox0\hbox{LIBERO-PRO Task~\cite{zhou2026liberopro}}\edef\tabTn{\the\wd0}\advance\dimen0 by \wd0
\setbox0\hbox{LIBERO-Plus~\cite{fei2026liberoplus}}\edef\tabPn{\the\wd0}\advance\dimen0 by \wd0
\setlength{\tabcolsep}{\dimexpr(\textwidth-\dimen0-18em-7.5pt-0.3pt)/30\relax}
\edef\tabTw{\the\dimexpr(\tabTn-2\tabcolsep)/2\relax}
\edef\tabPw{\the\dimexpr(\tabPn-2\tabcolsep)/2\relax}
\captionsetup{position=top,skip=4pt}
\caption{{Results on LIBERO, LIBERO-PRO and LIBERO-Plus. Gray: requires training ($\dagger$), not ranked. $\ast$: reported by the authors.}}
\label{tab:pro}
\begin{tabular}{lcw{c}{0pt}w{r}{2.25em}w{r}{2.25em}w{r}{\dimexpr 2.25em+2.5pt\relax}w{r}{2.25em}w{r}{\dimexpr 2.25em+2.5pt\relax}w{r}{2.25em}w{r}{\dimexpr 2.25em+2.5pt\relax}w{r}{2.25em}w{c}{0pt}w{r}{\tabTw}w{l}{\tabTw}w{c}{0pt}w{r}{\tabPw}w{l}{\tabPw}}
\toprule
 & LIBERO~\cite{liu2023libero} && \multicolumn{8}{c}{LIBERO-PRO Position~\cite{zhou2026liberopro}} &&
 \multicolumn{2}{c}{LIBERO-PRO Task~\cite{zhou2026liberopro}} && \multicolumn{2}{c}{LIBERO-Plus~\cite{fei2026liberoplus}}\\
\cmidrule(lr){2-2}\cmidrule(lr){4-11}\cmidrule(lr){13-14}\cmidrule(lr){16-17}
\multicolumn{1}{c}{Method} & Avg && \multicolumn{2}{c}{Object} & \multicolumn{2}{c}{Goal} & \multicolumn{2}{c}{Spatial} & \multicolumn{2}{c}{Avg} && \multicolumn{2}{c}{Avg} && \multicolumn{2}{c}{Layout}\\
 & \ml{Succ.} && \ml{Reach} & \ml{Succ.} & \ml{Reach} & \ml{Succ.} & \ml{Reach} & \ml{Succ.} & \ml{Reach} & \ml{Succ.} && \ml{Reach} & \ml{Succ.} && \ml{Reach} & \ml{Succ.}\\
\midrule
OpenVLA-OFT~\cite{kim2025oft} & \nb{96.8\rlap{$^{\ast}$}} && 0.2 & 0.6 & 29.6 & 0.0 & 85.4 & 27.6 & 38.4 & 9.4 && \nb{21.9} & \nb{9.7} && \nb{87.7} & \nb{72.6}\\
SmolVLA~\cite{shukor2025smolvla} & \nb{87.3\rlap{$^{\ast}$}} && 0.0 & 0.0 & 24.8 & 0.0 & 88.8 & 0.0 & 37.9 & 0.0 && \nb{14.5} & \nb{5.9} && \nb{88.5} & \nb{60.0}\\
TurboVLA & \nb{\textbf{97.7}\rlap{$^{\ast}$}} && 0.8 & 0.2 & 31.6 & 1.0 & 71.2 & 16.2 & 34.5 & 5.8 && \nb{17.3} & \nb{8.7} && \nb{88.1} & \nb{73.2}\\
\piO-FAST~\cite{pertsch2025fast} & \nb{85.5\rlap{$^{\ast}$}} && 0.0 & 0.0 & 48.6 & 4.6 & 90.0 & 29.0 & 46.2 & 11.2 && \nb{42.6} & \nb{27.3} && \nb{91.1} & \nb{74.3}\\
\piO~\cite{black2024pi0} & \nb{94.2\rlap{$^{\ast}$}} && 0.0 & 0.2 & 22.8 & 14.0 & 90.0 & 8.4 & 37.6 & 7.5 && \nb{33.0} & \nb{24.7} && \nb{90.4} & \nb{76.9}\\
\midrule
\trainedrow RSS\smash{$^\dagger$}~\cite{zhan2026rss} &\trainedrow \nb{96.6\rlap{$^{\ast}$}} &&\trainedrow 78.8 &\trainedrow 54.8 &\trainedrow 66.8 &\trainedrow 30.2 &\trainedrow 90.4 &\trainedrow 53.2 &\trainedrow 78.7 &\trainedrow 46.1 &&\trainedrow \nb{58.8} &\trainedrow \nb{37.1} &&\trainedrow \nb{96.2} &\trainedrow \nb{85.4}\\
\trainedrow APT\smash{$^\dagger$}~\cite{xu2026apt} &\trainedrow \nb{96.1\rlap{$^{\ast}$}} &&\trainedrow 12.4 &\trainedrow 10.6 &\trainedrow 64.0 &\trainedrow 7.2 &\trainedrow 88.4 &\trainedrow 38.6 &\trainedrow 54.9 &\trainedrow 18.8 &&\trainedrow \nb{50.1} &\trainedrow \nb{30.9} &&\trainedrow \nb{91.3} &\trainedrow \nb{76.7}\\
\midrule
\piH~\cite{black2025pi05} & \nb{96.7} && 22.0 & 16.6 & 75.8 & 33.2 & 90.0 & 45.2 & 62.6 & 31.7 && \nb{44.6} & \nb{29.6} && \nb{94.5} & \nb{\textbf{85.8}}\\
\piH\ + CAG~\cite{fang2026cag} & \nb{87.7} && 26.6 & 16.4 & 66.8 & 26.6 & 90.0 & 45.8 & 61.1 & 29.6 && \nb{47.7} & \nb{32.1} && \nb{\textbf{94.6}} & \nb{75.9}\\
\rowcolor[HTML]{EDEDED}
\piH\ + PSG (Ours) & \nb{95.2} && \textbf{52.6} & \textbf{40.8} & \textbf{87.0} & \textbf{40.8} & 90.0 & \textbf{52.4} & \textbf{76.5} & \textbf{44.7} && \nb{\textbf{50.4}} & \nb{\textbf{34.3}} && \nb{93.7} & \nb{82.5}\\
\bottomrule
\end{tabular}
\end{table*}

\section{Experiments}
\label{sec:experiments}

We evaluate whether PSG restores reaching after objects are relocated, while preserving performance on the trained tasks and improving robustness to instruction changes.

\providecommand{\trainedrow}{\color{black!55}\hypersetup{allcolors=black!55}}
\providecommand{\tbd}{\textcolor{gray!70}{--\,--}}
\providecommand{\nb}[1]{\makebox[2.25em][r]{#1}}
\providecommand{\ml}[1]{\makebox[2.25em][c]{\scriptsize\itshape #1}}
\begin{table}[t]
\footnotesize
\centering
\setlength\heavyrulewidth{0.25ex}
\renewcommand{\arraystretch}{1.0}
\setbox0\hbox{OpenVLA-OFT~\cite{kim2025oft}}\dimen0=\wd0
\setbox0\hbox{\piH\ + CAG~\cite{fang2026cag}}\ifdim\wd0>\dimen0 \dimen0=\wd0 \fi
\setbox0\hbox{\piH\ + PSG (Ours)}\ifdim\wd0>\dimen0 \dimen0=\wd0 \fi
\setlength{\tabcolsep}{\dimexpr(\columnwidth-\dimen0-13.5em-5pt-0.3pt)/14\relax}
\captionsetup{position=top,skip=4pt}
\caption{{Results by how far the target is moved.}}
\label{tab:distance}
\begin{tabular}{lw{r}{2.25em}w{r}{2.25em}w{r}{\dimexpr 2.25em+2.5pt\relax}w{r}{2.25em}w{r}{\dimexpr 2.25em+2.5pt\relax}w{r}{2.25em}}
\toprule
 & \multicolumn{6}{c}{LIBERO-Goal~\cite{liu2023libero}, target object moved (cm)}\\
\cmidrule(lr){2-7}
\multicolumn{1}{c}{Method} & \multicolumn{2}{c}{$<$\,20} & \multicolumn{2}{c}{20--30} & \multicolumn{2}{c}{$>$\,30}\\
 & \ml{Reach} & \ml{Succ.} & \ml{Reach} & \ml{Succ.} & \ml{Reach} & \ml{Succ.}\\
\midrule
OpenVLA-OFT~\cite{kim2025oft} & 5.0 & 0.0 & 2.5 & 0.0 & 0.0 & 0.0\\
SmolVLA~\cite{shukor2025smolvla} & 2.5 & 0.0 & 0.0 & 0.0 & 0.0 & 0.0\\
TurboVLA & 2.5 & 0.0 & 5.0 & 0.0 & 0.0 & 0.0\\
\piO-FAST~\cite{pertsch2025fast} & 80.0 & 17.5 & 50.0 & 0.0 & 3.0 & 0.0\\
\piO~\cite{black2024pi0} & 42.5 & 10.0 & 0.0 & 0.0 & 0.0 & 0.0\\
\midrule
\trainedrow RSS\smash{$^\dagger$}~\cite{zhan2026rss} &\trainedrow 100.0 &\trainedrow 52.5 &\trainedrow 100.0 &\trainedrow 90.0 &\trainedrow 46.0 &\trainedrow 20.0\\
\trainedrow APT\smash{$^\dagger$}~\cite{xu2026apt} &\trainedrow 100.0 &\trainedrow 57.5 &\trainedrow 67.5 &\trainedrow 10.0 &\trainedrow 12.0 &\trainedrow 4.0\\
\midrule
\piH~\cite{black2025pi05} & 100.0 & 47.5 & 55.0 & 57.5 & 24.0 & 6.0\\
\piH\ + CAG~\cite{fang2026cag} & 100.0 & 45.0 & 80.0 & 65.0 & 22.0 & \textbf{13.0}\\
\rowcolor[HTML]{EDEDED}
\piH\ + PSG (Ours) & 100.0 & \textbf{85.0} & \textbf{100.0} & \textbf{75.0} & \textbf{50.0} & \textbf{13.0}\\
\bottomrule
\end{tabular}
\end{table}

\subsection{Setup}
\label{sec:settings}

\noindent\textbf{Benchmarks.}
We evaluate on four benchmarks.
The original LIBERO~\cite{liu2023libero} measures success on the trained tasks, averaged over its four suites.
LIBERO-PRO~\cite{zhou2026liberopro} provides two perturbations on the Object, Goal, and Spatial suites: \emph{Position} relocates objects as in \cref{sec:mechanism}, while \emph{Task} changes the instruction without changing the scene.
We exclude its Long suite, which contains two consecutive subtasks.
We also evaluate the Layout perturbation of LIBERO-Plus~\cite{fei2026liberoplus} and the Spatial, Object, and OOD subsets of LIBERO-CF~\cite{fang2026cag}, which assigns alternative feasible instructions to LIBERO scenes.
For \cref{tab:distance}, we move the bowl, cream cheese, and wine bottle in LIBERO-Goal to six new positions each, leaving their training places empty.
Results are grouped by displacement from the training position: 4, 4, and 10 scenes fall below 20\,cm, at 20--30\,cm, and beyond 30\,cm, with 10 episodes for each scene (using start states 6--15).

\noindent\textbf{Compared methods.}
We compare the six released VLAs from \cref{sec:mechanism}, two training-based methods (RSS~\cite{zhan2026rss} and APT~\cite{xu2026apt}), and two training-free guidance methods for \piH\ (CAG~\cite{fang2026cag} and PSG).
SmolVLA executes 10 actions per query instead of its default 50 open-loop actions.
Methods requiring training are marked with $\dagger$.
CAG uses the instruction-free prediction as its reference, with the authors' recommended guidance scale of 1.5 for \piH.
On LIBERO-CF, we additionally include CAG-VA, which uses an author-trained language-free reference model ($\dagger$).

\providecommand{\trainedrow}{\color{black!55}\hypersetup{allcolors=black!55}}
\providecommand{\tbd}{\textcolor{gray!70}{--\,--}}
\providecommand{\nb}[1]{\makebox[2.25em][r]{#1}}
\providecommand{\ml}[1]{\makebox[2.25em][c]{\scriptsize\itshape #1}}
\begin{table}[t]
\footnotesize
\centering
\setlength\heavyrulewidth{0.25ex}
\renewcommand{\arraystretch}{1.0}
\setbox0\hbox{\piH\ + CAG~\cite{fang2026cag}}\dimen0=\wd0
\setbox0\hbox{\piH\ + CAG-VA$^\dagger$}\ifdim\wd0>\dimen0 \dimen0=\wd0 \fi
\setbox0\hbox{\piH\ + PSG (Ours)}\ifdim\wd0>\dimen0 \dimen0=\wd0 \fi
\setlength{\tabcolsep}{\dimexpr(\columnwidth-\dimen0-18em-7.5pt-0.3pt)/18\relax}
\captionsetup{position=top,skip=4pt}
\caption{{Comparison with CAG on LIBERO-CF.}}
\label{tab:cf}
\begin{tabular}{lw{r}{2.25em}w{r}{2.25em}w{r}{\dimexpr 2.25em+2.5pt\relax}w{r}{2.25em}w{r}{\dimexpr 2.25em+2.5pt\relax}w{r}{2.25em}w{r}{\dimexpr 2.25em+2.5pt\relax}w{r}{2.25em}}
\toprule
 & \multicolumn{8}{c}{LIBERO-CF~\cite{fang2026cag}}\\
\cmidrule(lr){2-9}
\multicolumn{1}{c}{Method} & \multicolumn{2}{c}{Spatial} & \multicolumn{2}{c}{Object} & \multicolumn{2}{c}{OOD} & \multicolumn{2}{c}{Avg}\\
 & \ml{Reach} & \ml{Succ.} & \ml{Reach} & \ml{Succ.} & \ml{Reach} & \ml{Succ.} & \ml{Reach} & \ml{Succ.}\\
\midrule
\piH~\cite{black2025pi05} & 30.3 & 24.7 & 7.4 & 5.6 & 15.5 & 6.5 & 17.7 & 12.3\\
\trainedrow \piH\ + CAG-VA\smash{$^\dagger$} &\trainedrow 41.7 &\trainedrow 32.5 &\trainedrow 22.6 &\trainedrow 13.8 &\trainedrow 27.9 &\trainedrow 10.7 &\trainedrow 30.7 &\trainedrow 19.0\\
\piH\ + CAG~\cite{fang2026cag} & \textbf{41.9} & \textbf{34.5} & 13.6 & 8.4 & 18.7 & \textbf{10.8} & 24.7 & 17.9\\
\rowcolor[HTML]{EDEDED}
\piH\ + PSG (Ours) & 38.8 & 30.9 & \textbf{28.4} & \textbf{12.8} & \textbf{25.5} & \textbf{10.8} & \textbf{30.9} & \textbf{18.2}\\
\bottomrule
\end{tabular}
\end{table}

\noindent\textbf{Metrics.}
\emph{Reach} is the percentage of episodes whose first gripper close (\cref{sec:mechanism}) lands on the requested target.
For tasks without grasping, such as opening a drawer or pushing a plate, we use the first contact instead.
\emph{Succ.} denotes task success.
For the original LIBERO results in \cref{tab:pro}, we evaluate \piH, CAG, and PSG ourselves; the other seven results are those reported by the authors ($\ast$).

\noindent\textbf{Implementation.}
PSG uses $a=5$ and $b=1$ throughout (\cref{sec:shortcut-removal}).
We select these values on 11 development tasks under the original, Position, and Task conditions, using start states 6--15, and keep them fixed for all evaluations.
These states are included among the 50 evaluation states.
Each original LIBERO, Position, and Task cell contains 500 episodes (10 tasks $\times$ 50 start states), evaluated with a single noise seed.
LIBERO-Plus Layout contains 1,525 episodes, and the three LIBERO-CF subsets contain 750, 500, and 750.
\piH\ predicts a chunk of 10 actions per replan and executes only the first five of them before replanning.

\subsection{Does PSG Restore the Reach of Moved Objects?}
\label{sec:main-results}

\noindent\textbf{PSG restores reaching at moved targets.}
On LIBERO-PRO Position, PSG improves \piH's average Reach from 62.6\% to 76.5\% (\cref{tab:pro}).
The largest gain is on Object, where Reach rises from 22.0\% to 52.6\% and task success from 16.6\% to 40.8\%.
For the milk, the released policy never reaches the moved object in 50 episodes, whereas PSG reaches it in 46.
These gains require neither retraining nor any externally supplied object positions.

\noindent\textbf{PSG approaches trained methods without training.}
RSS achieves 78.7\% average Reach and 46.1\% success under Position, compared with PSG's 76.5\% and 44.7\% (\cref{tab:pro}).
RSS leads on Object (78.8\% versus 52.6\% Reach) and Task (58.8\% versus 50.4\%), while PSG performs better on Goal (87.0\% versus 66.8\%).
For targets displaced beyond 30\,cm, PSG also achieves slightly higher Reach than RSS (50\% versus 46\%; \cref{tab:distance}), although RSS retains higher task success.
Thus, PSG approaches the average performance of a trained method, with substantial differences across tasks.

\begin{figure}[t]
\centering
\includegraphics[width=\linewidth]{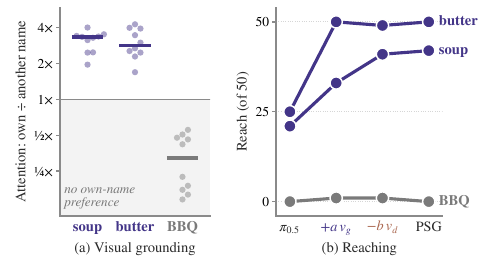}%
\caption{\textbf{PSG gains depend on existing name-to-object grounding.}
(a)~Name-token attention to each object's patches under its own name, relative to the strongest attention under another object's name. 
Values below $1\times$ indicate no preference for the object's own name.
(b)~First-close Reach for the same three objects under Position, with grounding amplification ($+a v_g$), direct-effect suppression ($-b v_d$), and both combined (PSG). Soup and butter improve, while BBQ sauce does not.}
\label{fig:found}
\end{figure}

\noindent\textbf{PSG extends reaching across displacement distances.}
On LIBERO-Goal, the released \piH\ reaches 100\%, 55\%, and 24\% of targets displaced by less than 20\,cm, 20--30\,cm, and more than 30\,cm (\cref{tab:distance}).
PSG maintains 100\% Reach within 20\,cm, raises the middle range to 100\%, and improves the farthest range to 50\%.
Even where both policies reach every target, PSG raises task success from 47.5\% to 85.0\%, so its benefit extends beyond the first reach.

\noindent\textbf{PSG largely preserves trained-task performance.}
On the original LIBERO suites, PSG achieves 95.2\% success, compared with 96.7\% for the released policy.
It also improves LIBERO-PRO Task, raising Reach from 44.6\% to 50.4\% and success from 29.6\% to 34.3\% (\cref{tab:pro}).
On LIBERO-Plus Layout, Reach decreases from 94.5\% to 93.7\%, and success from 85.8\% to 82.5\%.
APT~\cite{xu2026apt} raises average Task Reach to 50.1\%, but reduces Position Reach to 54.9\%, below the released policy's 62.6\%; PSG achieves 50.4\% and 76.5\%.
Thus, improving instruction following need not come at the expense of reaching relocated objects.

\noindent\textbf{Whole-instruction guidance can reinforce the shortcut.}
Unlike PSG, CAG contrasts predictions with and without the entire instruction, combining the opposing pathway effects identified in \cref{sec:cut}.
CAG lowers average Position Reach from 62.6\% to 61.1\% and original LIBERO success from 96.7\% to 87.7\% (\cref{tab:pro}).
Although CAG improves reaching for the orange juice, it reduces Reach for the alphabet soup from 21 to 5 of 50 episodes.
Most first closes instead land on the butter, which occupies the soup's training place.
Thus, amplifying the whole instruction can also reinforce the instruction shortcut.
On LIBERO-CF, PSG improves Object Reach from 7.4\% to 28.4\%, compared with CAG's 13.6\% (\cref{tab:cf}).
CAG performs slightly better on Spatial (41.9\% versus 38.8\%), while PSG achieves higher average Reach (30.9\% versus 24.7\%).
Without training, PSG matches the average Reach of the trained CAG-VA (30.7\%), with slightly lower success (18.2\% versus 19.0\%).

\begin{figure}[t]
\centering
\includegraphics[width=\linewidth]{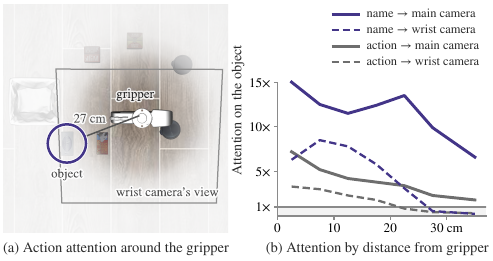}%
\caption{\textbf{Name attention tracks the object, while action attention concentrates near the gripper.}
(a)~Top view of the gripper near the training place, with a local attention halo and the wrist-camera field of view. 
Here, gripper-to-object distance approximately reflects target displacement.
(b)~Name- and action-token attention to the requested object as a function of gripper-to-object distance, separated by camera view and normalized by image-average attention. Main-camera name attention remains high, while action attention decreases with distance.}
\label{fig:halo}
\end{figure}

\noindent\textbf{Both pathway corrections can improve reaching.}
For soup and butter, amplifying the grounding effect or suppressing the direct effect improves reaching (\cref{fig:found}b).
Combining them preserves or further improves these gains.
This supports adjusting the two pathway effects separately rather than amplifying their whole-instruction contrast.

\subsection{Discussion and Implications}
\label{sec:where}

The gains from PSG invite a different reading of relocation failures: returning to a familiar place need not mean that the policy lacks the ability to act elsewhere.
It may instead reveal a gap between what the policy has learned and the behavior that it normally generates on its own.

\begin{figure}[t]
\centering
\includegraphics[width=\linewidth]{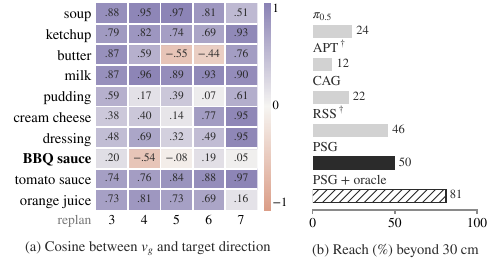}%
\caption{\textbf{The grounding effect points toward the target.}
(a)~Cosine between the translational grounding effect $v_g$ and the gripper-to-target direction across replans for ten objects.
(b)~Reach beyond 30\,cm; an oracle target-direction push in place of $v_g$ raises it from 50\% to 81\%.}
\label{fig:replan}
\end{figure}

\noindent\textbf{Similar failures can conceal different grounding states.}
In \piO, name attention does not follow the moved object, while in \piH\ it does, even though both mostly return to training locations (\cref{sec:mechanism}).
This difference also appears across objects within \piH.
For soup and butter, name attention identifies the moved object and PSG improves reaching.
For BBQ sauce, name attention shows little preference for the requested object (\cref{fig:found}a), and the grounding effect lacks a reliable target direction (\cref{fig:replan}a).
Neither pathway correction improves its reaching (\cref{fig:found}b).
Thus, similar reaching failures can accompany either weak or retained visual grounding.
The behavior alone does not tell us which.

\noindent\textbf{Why can grounding and the shortcut coexist?}
The direct pathway is not inherently a shortcut; its effect depends on what the policy has learned.
In familiar demonstrations, locating the named object and recalling its usual position lead to the same reach.
Training success therefore provides little pressure to distinguish these behaviors.
Indeed, cutting the direct pathway preserves 199 of 200 reaches in the original scenes, yet increases reaches to moved objects from 29 to 64 of 200 (\cref{sec:cut}).
Relocation exposes a conflict that familiar demonstrations do not require the policy to resolve.

\noindent\textbf{Visual feedback may refine, not redirect, a movement.}
Prior work suggests that instructions can retrieve familiar trajectories while vision adapts their execution~\cite{li2025vlasconfined,chen2026instructions}.
Our observations offer a possible spatial explanation.
Name attention through the main camera remains focused on distant objects, while action-token attention to the object decreases with gripper-to-object distance, particularly through the wrist camera (\cref{fig:halo}).
Such locality may be useful for a policy that generates short action chunks and frequently replans: nearby visual feedback can support grasping and other local adjustments.
But when an instruction shortcut has already drawn the arm toward a familiar place, local visual feedback may refine that movement without redirecting the arm toward the moved object's new location.

The distance-dependent behavior is consistent with this interpretation.
Even \piO\ and \piO-FAST, whose name attention does not reliably follow moved objects, reach some nearby relocations but almost none beyond 30\,cm (\cref{tab:distance}).
These observations suggest that local visual adaptation may work even without reliable name-to-object grounding, while larger relocations demand a stronger influence from the current target.

\noindent\textbf{Grounding already points toward the target.}
For most of the ten tested objects, the grounding effect $v_g$ already points toward the target, the direction an oracle would supply (\cref{fig:replan}a).
For LIBERO-Goal targets displaced beyond 30\,cm, an oracle push along this direction raises Reach from 50\% to 81\% (\cref{fig:replan}b).
The direction that PSG amplifies is thus the right one, and strengthening the policy's own target signal leaves clear room for further gains.

Together, these analyses show that returning to the training place can arise for different reasons.
A policy may fail to locate the moved object, locate it but follow an instruction shortcut, or produce a target-directed grounding effect that is insufficient to redirect reaching over larger displacements.
Crucially, such failures need not imply missing action skills: without updating the policy's parameters, PSG raises task success on LIBERO-PRO Object under Position from 16.6\% to 40.8\% (\cref{tab:pro}).
Where grounding is retained, the challenge is therefore not only to learn new skills, but to make the grounded target more reliably determine where existing skills are executed.
\section{Conclusion}
\label{sec:conclusion}

We investigated why VLAs return to familiar locations when objects are moved. 
Across six models, we identified scene and instruction shortcuts. 
In \piH, visual grounding survives relocation and already guides action, but is often overridden by an instruction shortcut through the direct pathway. 
Pathway-Separated Guidance (PSG) separates these opposing effects, improving reaching and task success without retraining or external target locations.

These findings suggest that a policy's default behavior need not reflect the full extent of its learned capabilities. 
Some skills can already be executed beyond their demonstrated locations, even when the policy fails to reach them on its own. 
The challenge is therefore not only to preserve visual grounding or learn new skills, but to let the objects a policy sees determine where its existing skills are used.

\section*{LLM Usage Statement}
We used Claude Code (Anthropic), and Codex and ChatGPT (OpenAI), in this work.
\textbf{Research.} We discussed research ideas and progress with these tools.
Claude Code implemented the method and evaluation code from our designs, ran the evaluations with our approval, aggregated the results, and wrote the plotting code for the figures.
\textbf{Writing.} We drafted the main line of each section and the content of each paragraph; ChatGPT and Claude Code rendered the drafts into English and suggested edits, and we revised the text sentence by sentence.
\textbf{Literature.} Claude Code assisted the literature search and screened the reference list; we verified that every cited work exists and imported each BibTeX entry by hand from the publisher's site or arXiv.
We formulated the research question, designed and approved the experiments, interpreted the results, and compiled the tables.
We reviewed every sentence of the final manuscript and take full responsibility for its content.
{\small
\bibliographystyle{ieeenat_fullname}
\bibliography{references}
}
\clearpage
\ifdefined\linenumbers\edef\suppLNoffset{\arabic{linenumber}}\renewcommand\thelinenumber{\fillzeros[3]{\numexpr\value{linenumber}-\suppLNoffset\relax}}\fi
\renewcommand{\thefigure}{S\arabic{figure}}\setcounter{figure}{0}
\renewcommand{\thetable}{S\arabic{table}}\setcounter{table}{0}
\extrafloats{100}
\renewcommand{\topfraction}{0.95}\renewcommand{\dbltopfraction}{0.95}\renewcommand{\bottomfraction}{0.8}
\renewcommand{\textfraction}{0.05}\renewcommand{\floatpagefraction}{0.5}\renewcommand{\dblfloatpagefraction}{0.5}
\setcounter{topnumber}{4}\setcounter{dbltopnumber}{4}\setcounter{bottomnumber}{2}\setcounter{totalnumber}{6}
\providecommand{\ml}[1]{\makebox[2.25em][c]{\scriptsize\itshape #1}}
\newcommand{\suppfloat}[1]{\InputIfFileExists{sec/suppl/#1/float.tex}{}{\par\noindent\fbox{\parbox{0.95\linewidth}{\small #1: not in place yet.}}\par}}
\newcommand{\supptext}[1]{\InputIfFileExists{sec/suppl/text/#1.tex}{}{\par\noindent{\small\color{gray}[#1: text to come.]}\par}}
\newcommand{\partfloats}{\par\suppressfloats[t]}
\makeatletter
\newcommand{\suppblock}[3]{\begingroup\def\supplw{#2}%
  \renewenvironment{table*}[1][]{\def\@captype{table}\par\noindent\begin{minipage}{\supplw}\centering}{\end{minipage}\par#3}%
  \renewenvironment{table}[1][]{\def\@captype{table}\par\noindent\begin{minipage}{\supplw}\centering}{\end{minipage}\par#3}%
  \renewenvironment{figure*}[1][]{\def\@captype{figure}\par\noindent\begin{minipage}{\supplw}\centering}{\end{minipage}\par#3}%
  \renewenvironment{figure}[1][]{\def\@captype{figure}\par\noindent\begin{minipage}{\supplw}\centering}{\end{minipage}\par#3}%
  \input{sec/suppl/#1/float.tex}\endgroup}
\makeatother
\newenvironment{supppage}{\begin{figure*}[p]\centering}{\vspace{-\dblfloatsep}\end{figure*}}
\setlength{\dbltextfloatsep}{12pt plus 2pt minus 4pt}
\newenvironment{supptop}{\begin{figure*}[t]\centering}{\vspace{-\dblfloatsep}\end{figure*}}
\newcommand{\suppfull}[1]{\suppblock{#1}{\textwidth}{\vspace{\dblfloatsep}}}
\newcommand{\suppw}[2]{\suppblock{#1}{#2}{\vspace{\dblfloatsep}}}
\newcommand{\suppside}[2]{\noindent\begin{minipage}[t]{0.485\textwidth}\vspace{0pt}\suppblock{#1}{\linewidth}{}\end{minipage}\hfill\begin{minipage}[t]{0.485\textwidth}\vspace{0pt}\suppblock{#2}{\linewidth}{}\end{minipage}\par\vspace{\dblfloatsep}}
\makeatletter
\setlength{\@fptop}{0pt}\setlength{\@fpbot}{0pt plus 1fil}\setlength{\@dblfptop}{0pt}\setlength{\@dblfpbot}{0pt plus 1fil}
\makeatother
\providecommand{\mlw}[1]{\makebox[2.6em][c]{\scriptsize\itshape #1}}
\providecommand{\mlh}[1]{{\scriptsize\itshape\renewcommand{\arraystretch}{0.8}\begin{tabular}[b]{@{}c@{}}#1\end{tabular}}}
\newcommand{\suppcol}[1]{\par\vspace{8pt}\ifdefined\nolinenumbers\begin{nolinenumbers}\fi\suppblock{#1}{\columnwidth}{\vspace{4pt}}\ifdefined\nolinenumbers\end{nolinenumbers}\fi}
\newcommand{\suppcolfloat}[1]{\begin{figure}[tp]\suppblock{#1}{\columnwidth}{}\end{figure}}
\newcommand{\suppfoot}[1]{\par\vfill\ifdefined\nolinenumbers\begin{nolinenumbers}\fi\noindent\makebox[\columnwidth][l]{\begin{minipage}[b]{\textwidth}\suppblock{#1}{\textwidth}{}\end{minipage}}\par\ifdefined\nolinenumbers\end{nolinenumbers}\fi\newpage}
\newcommand{\suppalignpair}[2]{\par\vspace{8pt}\edef\supptextht{\the\pagetotal}\suppblock{#1}{\columnwidth}{}\newpage\vspace*{\supptextht}\suppblock{#2}{\columnwidth}{}\par}
\raggedbottom
\appendix
\newpage
\twocolumn[{\centering\Large\textbf{\thetitle}\\
\vspace{0.5em}Supplementary Material\\
\vspace{1.0em}\normalsize\begin{minipage}{\textwidth}\setlength{\parindent}{0pt}\noindent\textbf{Overview.}
Parts A to C continue the main paper's Secs.~3.1--3.2, 3.3 and 4--5: the two shortcuts and name attention across models (\cref{supp:field}), the pathway cuts (\cref{supp:cuts}), and PSG (\cref{supp:psg}).
\Cref{supp:protocol} gives the protocol, the algorithm, the settings and the per-task tables behind the main paper's Tabs.~1--3, and \cref{supp:limits} the limitations.
In Parts A to C, the bold sentence that opens a paragraph states its result.
[Overview: text to come.]}\par}\end{minipage}\par\vspace{8pt}\suppblock{A3}{0.88\textwidth}{\vspace{2pt}}}]

\noindent\textbf{Terms.}
Each LIBERO-Object task places its requested object at one of two fixed positions: the one in its demonstrations is its \emph{training place}, and the other, which belongs to other tasks, is the \emph{other training place}.
\emph{Position} exchanges the requested object with another object and keeps the instruction, so another object stands at the training place; \emph{Task-paired} keeps the scene and gives the training instruction of another object in it (\cref{supp:A3}).
The \emph{first close} is the first policy step at which the gripper command changes from open to closed, credited to the nearest object or fixture.
A \emph{reach} is a first close on the requested object, and \emph{Reach} is the share of episodes with a reach.
Where the exchange puts the requested object on the other training place, a close on it could reflect a shortcut rather than grounding: the \emph{conservative criterion} does not count reaches in these tasks, which the tables mark with $\ddagger$, and the \emph{looser criterion} counts them.
\emph{Name attention} is the attention from the requested object's name tokens to the main-camera image patches of an object, the requested one unless a table names another, relative to the image average.
\Cref{supp:protocol-text} gives the full protocol.
[Terms: text to come.]}\par}

\section{Shortcuts and Name Attention}
\label{supp:field}
The main paper's Sec.~3 rests on where six VLAs first close the gripper and on where the object's name attends.
This part gives both by model and object, and then measures where the name lands in the image tokens of \piH\ and \piO\ (\cref{supp:name-in-image}).
[Aopen: text to come.]}\par}
\subsection{The Two Shortcuts in Six VLAs}
\label{supp:shortcuts}
\begin{supptop}
\suppfull{A1}
\end{supptop}
\noindent\textbf{Three VLAs follow the scene, and the $\pi$ family follows the instruction.}
With the scene unchanged and another object requested (Task-paired), OpenVLA-OFT and TurboVLA first close at the place of the scene's original task in every episode, and SmolVLA in 167 of 200 (\cref{supp:A1}).
\piO, \piO-FAST and \piH\ instead first close at the requested object's training place, in 197, 191 and 157 of 200 episodes.
\Cref{supp:A3} draws the three places a first close can land at.

\noindent\textbf{Under Position, only \piH\ reaches the moved object.}
With the requested object exchanged with another (Position), the five other VLAs first close at its training place in 796 of 800 episodes and never on the object (\cref{supp:A1}).
\piH\ first closes there in 121 of 200 episodes and on the moved object in 42, of which the conservative criterion counts 28 and leaves out the 14 where the object stands on the other training place.
With the checkpoints of the main paper's Tab.~1 and all ten Object tasks, the five other models first close at the training place in 466 to 496 of 500 episodes; \cref{supp:D1b} gives their Reach per task.
[A0a: text to come.]}\par}
\subsection{Name Attention across Models and Objects}
\label{supp:name-attention}
\begin{supptop}
\suppfull{A1n}
\end{supptop}
\noindent\textbf{Only \piH's name attention averages well above the image average, and only under the object's own name.}
Averaged over nine moved objects, it is 6.38 times the image average on the object under the object's own name, and 1.30 times when another object of the scene is named (\cref{supp:A1n}(a)).
For OpenVLA-OFT, SmolVLA, \piO-FAST and \piO, the same average stays between 0.79 and 1.50 times under the object's own name.
In the Task-paired scenes, where only the request changes, \piH's name attends to the requested object at 4.50 times the image average, to the object at its training place at 2.30 times, and to the place of the scene's original task at 1.00; \piO's name stays below the image average at all three (\cref{supp:A1n}(b)).

\noindent\textbf{The shift repeats object by object, and the \piH\ base model already shows it.}
For each of the eight other objects, \piH's name attention leaves the training place when the object moves, keeping at most 2\% of the name's attention there, and lands on the new places with 4\% to 39\% (\cref{supp:A4}).
Before LIBERO fine-tuning, the \piH\ base model's name attention on the moved object is above that of the \piO\ and \piO-FAST base models for every one of the nine objects; after fine-tuning, \piH's is the highest of the four policies for eight of them (\cref{supp:A5}).
[A0b: text to come.]}\par}
\begin{supppage}
\setlength{\dblfloatsep}{6pt plus 1pt}
\suppfull{A4}
\suppw{A5}{0.76\textwidth}
\end{supppage}
\subsection{The Name in the Image Tokens}
\label{supp:name-in-image}
\suppcolfloat{Gmap}
Name attention is read at the instruction tokens, while the grounding pathway of the main paper's Sec.~3.3 runs through the image tokens that read the instruction.
We therefore measure the image tokens themselves: in the first frame of a Position episode, we replace only the requested object's name with the name of another object in view and record how much the keys and values of each main-camera image token change (\cref{supp:Gmap}; values in \cref{supp:Gkv}).

\noindent\textbf{In \piH, a change of the name lands on the two named objects; in \piO, it lands elsewhere.}
In \piH, the image tokens on the patches of the requested object and of the substitute change more than those on the patches that hold neither object, in every one of the ten tasks and in every layer band, by a median factor of 1.6 to 1.9 in the keys.
In \piO, they change less than those patches in every task and band, by a median factor of 0.4 to 0.5 (\cref{supp:Gkv} gives every band, for keys and values).
\Cref{supp:Gmap} shows the change per patch in one scene.
The image tokens' read of the instruction, which the grounding cut removes, thus carries the name to the named objects' tokens more than to the rest of the image in \piH, and less in \piO.
[A0c: text to come.]}\par}
\section{Pathway Cuts}
\label{supp:cuts}
The main paper's Sec.~3.3 cuts two pathways from instruction to action.
This part shows what each cut removes, how the two cuts move \piH's first closes, and what the direct pathway carries.
[Bopen: text to come.]}\par}
\subsection{What a Pathway Cut Removes}
\label{supp:mask}
\begin{supptop}
\suppw{B5}{0.9\textwidth}
\end{supptop}
Each cut is an attention mask applied before the softmax in every layer, head and denoising step, so a masked query--key pair receives zero weight.
Because positions exchange information only through attention, the instruction reaches the action tokens by two routes, directly or through image tokens that have read it, and each cut closes one of them.

\noindent\textbf{Each cut removes exactly its route.}
With no cut, the masked model reproduces the released \piH\ exactly at the first query of every episode.
Under the grounding cut, reordering the instruction's words leaves the image tokens' keys and values unchanged in every layer, and exchanging only the requested object's name changes nothing in the keys and values of the object's patches.
Under the direct cut, replacing the instruction's keys and values leaves the action chunk unchanged, while without the cut the same change alters it in every episode (\cref{supp:B0chk} lists every check).
\Cref{supp:B5} draws both masks and the name-only variants, which restrict either cut to the requested object's name tokens $N_r$.
[B0: text to come.]}\par}
\subsection{Opposite Effects of the Two Cuts}
\label{supp:cut-results}
\begin{supptop}
\suppfull{B2}
\end{supptop}
\noindent\textbf{The two cuts move \piH's first closes in opposite directions.}
Under Position, the grounding cut lowers \piH's reaches from 29 to 0 of 200 and the direct cut raises them to 64 (conservative criterion; \cref{supp:B2}).
Paired episode by episode, the direct cut moves first closes from the training place onto the moved object, while under the grounding cut none of the 29 reaches remains (\cref{supp:B1}(a); \cref{supp:B7} shows one episode).
The direct cut also raises task success from 15 to 81 of 200 episodes (\cref{supp:B2}).
With both cuts, the reaches are 4 of 200: the reaches that the direct cut adds are not there without the grounding pathway.
The grounding cut lowers the action tokens' attention to the object's patches to 0.54 of its value with no cut in the median episode, and the direct cut leaves it at 1.04 (\cref{supp:B6}(b)).

\noindent\textbf{In \piO, cutting the direct pathway does not release the hand.}
Under the direct cut, \piO\ first closes at the training place in 198 of 200 episodes, against 197 with no cut, and never on the moved object (\cref{supp:B2}); neither cut changes its action tokens' attention to the object's patches, at 1.01 and 0.97 of the value with no cut (\cref{supp:B6}(b)).
Under the grounding cut, 72 of \piO's 200 first closes land at the other training place, 31 of them on the requested object in the two tasks where it stands there; the conservative criterion does not count these and gives no reach in the 200 episodes and 2 in the 250 of the repeat.
This agrees with \cref{supp:name-in-image}: in \piO, the image tokens' read of the name changes the named objects' tokens less than the rest of the image, so its grounding pathway does not single out the object.

\noindent\textbf{The grounding pathway's effect sits on the object's name.}
Restricted to the requested object's name tokens, the grounding cut still lowers the reaches to 0, as the full cut does, while the direct cut restricted to the same tokens leaves them at 30, against 29 with no cut (\cref{supp:B2}).
PSG's masks follow from this: on the grounding pathway they deny the main camera only the name tokens, and on the direct pathway the whole instruction (main paper, Sec.~4.1).

\noindent\textbf{The result repeats on 250 episodes per cut.}
With start states 0 to 49, the grounding cut lowers the reaches from 35 to 0 and the direct cut raises them to 71; task success goes from 37 to 10 and to 95 (lower block of \cref{supp:B2}).

\noindent\textbf{In the original scenes, the hand reaches the object with either pathway cut.}
With the direct pathway cut, the first close lands on the requested object in 249 of 250 episodes; with the grounding pathway cut, in 213 of 250 (\cref{supp:B2}), and in 47 to 50 of 50 in four of the five tasks.
Where the object stands at its training place, either pathway leads the hand to it; moving the object sets the two against each other.
[B2lead: text to come.]}\par}
\begin{supppage}
\setlength{\dblfloatsep}{3pt plus 1pt}
\suppfull{B1B6}
\suppw{B7}{0.86\textwidth}
\suppfull{G1}
\end{supppage}
\subsection{What the Direct Pathway Carries}
\label{supp:exchange}
To see what the direct pathway carries, we let the action tokens read the scene's instruction, a second object's instruction or nothing, while the image tokens read the scene's instruction or nothing (\cref{supp:G1}).
We use two Position tasks, the alphabet soup and the orange juice, with 50 episodes per cell; each second object was trained at the suite's other training place.

\noindent\textbf{Another instruction in the direct pathway releases the hand as the cut does.}
On the alphabet soup, the hand first closes at the soup's training place in 29 of 50 episodes when the action tokens read the scene's instruction, and in 9 when they read the second object's instruction or nothing; reaches rise from 21 to 41 (\cref{supp:G1}).
With nothing in the image tokens, the scene's instruction read by the action tokens still sends 29 of 50 first closes to the soup's training place, and none go there without it.

\noindent\textbf{A second object's instruction can bring its own training place.}
On the orange juice, with the scene's instruction in the image tokens, reading the second object's instruction sends 16 of 50 first closes to that object's training place, and reading the scene's instruction or nothing sends none (\cref{supp:G1}).
With nothing in the image tokens, it sends all 50 there, against 30 when nothing is read.

In each task, one instruction draws first closes to its own training place in every cell of \cref{supp:G1} where the action tokens read it: the scene's on the alphabet soup, the second object's on the orange juice.
The direct pathway thus carries the instruction shortcut of the main paper's Sec.~3.3.
[B3: text to come.]}\par}
\section{Pathway-Separated Guidance}
\label{supp:psg}
This part asks whether PSG's gain on LIBERO-PRO Position Object in the main paper's Tab.~1 is robust, what each of the two corrections contributes, and where the gain comes from.
[Copen: text to come.]}\par}
\renewcommand{\floatpagefraction}{0.9}
\suppcolfloat{C5}
\subsection{Robustness of the Gain}
\label{supp:psg-robust}
\suppcolfloat{C4p2}
\begin{supptop}
\suppfull{C4p2fig}
\end{supptop}
\noindent\textbf{The gain holds under the conservative criterion.}
On the five Position Object tasks in which the requested object does not stand on the other training place, PSG raises Reach from 14.0\% to 46.4\% and success from 8.8\% to 25.2\% (\cref{supp:C5}).
On the orange juice, Reach rises from 28\% to 100\% (\cref{supp:D1b}).

\noindent\textbf{The gain holds on episodes that the selection of the weights did not see.}
The two weights were selected on eleven development tasks, where $a=5$ gives 139 of 200 reaches under Position, against 119 at $a=2$ and 95 for the released policy (\cref{supp:protocol-text}).
On the episodes that share neither a task nor a start state with them, PSG raises Position Object Reach from 18.2\% to 42.5\% and success from 13.9\% to 28.2\% (\cref{supp:C5}).

\noindent\textbf{The gain holds at every tested weight $a$.}
With $b=1$, we vary $a$ on the ten Position Object tasks, 20 episodes per task, on start states the selection did not use (\cref{supp:C4p2}).
The sampler reaches the object in 98 to 107 of 200 episodes at $a=1$, 2, 5 and 8, against 42 for the released policy.

\noindent\textbf{The gain repeats with a second noise seed, episode by episode.}
With noise seed 17 in place of 7 on the same 500 Position Object episodes, PSG raises Reach from 22.6\% to 51.4\% and success from 17.2\% to 39.2\%, against 22.0\% to 52.6\% and 16.6\% to 40.8\% in the main paper's Tab.~1 (\cref{supp:C6}).
Paired by task and start state, PSG reaches the object in 153 episodes in which the released policy does not and in none the reverse at seed 7, and in 147 against 3 at seed 17.

\noindent\textbf{After the reach, PSG completes the task at least as often as the released policy on each suite.}
Of the episodes with a reach, the released policy succeeds in 75\% and PSG in 78\% on Position Object (83 of 110 and 204 of 263), in 44\% and 46\% on Goal (166 of 379 and 199 of 435), and in 50\% and 58\% on Spatial (226 of 450 and 262 of 450).
In six of the twenty Goal and Spatial tasks, no method succeeds in a single episode, trained or not, and in five of them the released policy and PSG reach the target in 60\% to 100\% of the episodes (\cref{supp:D1b}).
These tasks hold 149 of the 236 Goal episodes in which PSG reaches the target and does not succeed.
[C0a: text to come.]}\par}
\suppcolfloat{C6}
\subsection{The Two Corrections}
\label{supp:psg-ablation}
\noindent\textbf{Each correction alone raises Reach over the ten tasks, and the two together give the most successes.}
Amplifying only the grounding effect ($a=5$, $b=0$) gives 89 of 200 reaches and removing only the direct effect ($a=0$, $b=1$) gives 75, against 42 for the released policy and 105 for PSG (\cref{supp:C4p2}).
Successes are 52 and 58 with one correction, and 84 with both.

\noindent\textbf{Subtracting the grounding effect sends the hand back to the training place.}
With $a=-1$ and $b=0$, the sampler integrates $v[I_m \nrightarrow N]$ on the motion dimensions: the policy's motion runs under PSG's own grounding mask.
Reaches on the three development tasks of LIBERO-Object then fall from 18 to 2 of 60 (\cref{supp:C4p2}): there, the effect that PSG amplifies is the one that carries the released policy's reaches.
The first close then lands on the object standing at the requested object's training place in 19 or 20 of 20 episodes in six of the ten tasks (\cref{supp:C4p2fig}).
[C0b: text to come.]}\par}
\subsection{Where the Gain Comes From}
\label{supp:psg-source}
\suppcolfloat{C3}
\noindent\textbf{PSG adds Reach where the name finds the object.}
The own-name ratio is the object's name attention under its own name over the strongest under another object's name.
It is above 1 in all ten frames for the alphabet soup, the butter, the chocolate pudding and the orange juice, and PSG adds 21 to 36 reaches of 50 on their tasks; it is below 1 in all ten frames for the BBQ sauce, where PSG adds none (\cref{supp:C3}).
The same holds across models: on \piO, whose name attention stays near the image average (\cref{supp:A1n}), the same sampler leaves Reach at 6 of 500 Position Object episodes, against 0 for the released \piO, and no episode succeeds (\cref{supp:C6}).

\noindent\textbf{On the Object grid, PSG gains most at 20 to 30\,cm.}
The Object grid places each of ten LIBERO-Object objects at its training place and at six new places (\cref{supp:C2}).
At the places 20 to 30\,cm from the training place, PSG raises the reaches from 16 to 67 of 143 episodes, with more reaches than the released policy at 12 of the 15 places and fewer at none.
Within 20\,cm, it raises them from 179 to 196 of 200.
Moved only 12\,cm, the BBQ sauce is reached as well: PSG raises its Reach from 50\% to 100\%.

\noindent\textbf{Along the true direction, the same correction reaches farther.}
On the Goal scenes with the target moved more than 30\,cm, a push along the true direction in place of $v_g$, with the direct effect suppressed as in PSG, reaches the target in 81\% of episodes, against 50\% with PSG (main paper, Fig.~7(b)).
A stronger target signal in the same sampler thus leaves room for further gains.
[C0c: text to come.]}\par}
\section{Experimental Details}
\label{supp:protocol}
\subsection{Protocol}
\label{supp:protocol-text}
\noindent\textbf{Places and scenes.}
The places and the two perturbations are defined at the head of this document.
Task-paired uses the scenes of five LIBERO-Object tasks (0, 3, 6, 8 and 9); these are the scenes in which the requested object's training place differs from the place of the scene's original task and the requested object is not at its own training place.
A first close can then land at three distinct places: the requested object, its training place and the place of the scene's original task (\cref{supp:A3}).

\noindent\textbf{First close and Reach.}
A first close is credited to the object or fixture whose geometry lies nearest the gripper; in the experiments of the main paper's Secs.~3.1 and~3.3 it is credited to the object or basket nearest the end effector in the horizontal plane.
Tasks without grasping use the first contact.
Success is the benchmark's own task check and does not depend on the first close, so an episode whose first close lands elsewhere can still succeed.
Over the 30,147 reaches that a first close decides in the rows of the main paper's tables, the gripper lies a median of 1.6 to 2.3\,cm from the nearest geometry center of the credited object in each row (1.6\,cm for \piH\ and 1.8\,cm with PSG), and more than 15\,cm in 71 of them (0.2\%).
In the cut experiments, the conservative criterion does not count first closes on the requested object in the two tasks where the exchange puts it on the other training place (the BBQ sauce's and the butter's), and all 200 episodes stay in the denominator.
In the per-task tables behind the main paper's Tab.~1, its line leaves the five tasks so affected out of both numerator and denominator (\cref{supp:D1b}).

\noindent\textbf{Attention numbers.}
Most attention values are the attention on a region's patches divided by the camera's mean per patch; \cref{supp:D6} gives the statistic, tokens, layers and frames behind each attention figure and table.
Name attention across training stages (\cref{supp:A5}) and of the LIBERO policies (\cref{supp:A1n}(a)) is read on start frames with the object moved and the wrist image blank, every touched patch counting equally, as a geometric mean over nine objects.
Name attention in the Task-paired scenes (\cref{supp:A1n}(b)) is read on the first policy frames with the real wrist image and pixel-weighted patches, as one median over 50 frames.
The two protocols differ in wrist image and patch weighting, so values are compared within a panel.

\noindent\textbf{Settings.}
\Cref{supp:D3runs} gives the models, scenes, start states, noise seeds and episodes behind every figure and table that reports a measurement, \cref{supp:D2} the test sets of the main paper's tables, and \cref{supp:D3} the checkpoints; \piH\ and \piO\ are the released \texttt{pi05\_libero} and \texttt{pi0\_libero}.
On LIBERO, LIBERO-PRO and LIBERO-CF the main paper's tables use start states 0 to 49, and all of them use noise seed 7 (\cref{supp:D2}).
The cuts of the main paper's Sec.~3.3 and of \cref{supp:cuts} run \piH\ and \piO\ on five Object tasks, under Position, and in the original scenes with no cut and the direct cut, with start states 0 to 19 and noise seeds 7 and 17, 200 episodes per condition; the lower block of \cref{supp:B2} repeats them under Position, and for \piH\ in the original scenes, with start states 0 to 49 and seed 7, 250 episodes per condition.
Their no-cut condition repeats the Position episodes of \cref{supp:A1} and gives 29 and 43 reaches by the conservative and the looser criterion, against 28 and 42 there.
In \cref{supp:exchange}, the second object is chosen so that the two objects and their two training places lie at least 15\,cm apart.
The second-seed rows of \cref{supp:C6} change only the sampler's noise seed, to 17; they ran on another machine than the tables' rows, and at seed 7 the two machines give the same Reach in all 400 paired episodes that both ran.
On \piO, PSG runs with \piO's own inference settings, and its direct mask also denies the state token the instruction, as the direct cut does.

\noindent\textbf{Selection of the weights.}
PSG's two weights were selected on eleven development tasks, tasks 0, 6 and 8 of LIBERO-Object, 0, 6 and 9 of LIBERO-Goal, 1, 3 and 4 of LIBERO-Spatial and 0 and 2 of LIBERO-Long, each under the original scenes, Position and Task, with start states 6 to 15 and noise seeds 7 and 17.
Where the selection counts Reach (not on the drawer task of LIBERO-Goal), $a=5$ gives 139 of 200 reaches under Position and 140 of 200 under Task, against 119 and 118 at $a=2$ and 95 and 100 for the released policy.
In the original scenes, where the selection used success, the three settings succeed in 213, 211 and 212 of 220 episodes.

\noindent\textbf{Checks and the name exchange.}
At the first query of every episode, with no cut, the action chunk and the executed actions are compared with the released policy's; under the grounding cut, the image tokens' keys and values in every layer are compared between the instruction and its words reordered; under the direct cut, the action chunk is compared with the one obtained after the instruction tokens' keys and values in every layer are set to other numbers ($x\to -x-1$).
A check must be 0, and its control, the same difference with nothing cut, must be above 0 (\cref{supp:B0chk}).
Under the grounding cut, the exchange of only an object's name is tested on frames of eight objects at 29 places in LIBERO-Goal and LIBERO-Object scenes, at replans 0 and 4, and changes none of the object's main-camera keys and values.
The changes of \cref{supp:Gkv} are read with nothing cut, on the Position first frame of the ten LIBERO-PRO Object tasks at start states 6 to 15 (100 frames): the requested object's name is replaced with that of each of the five other objects in view (500 pairs per model), and a patch counts for an object if it holds a pixel of it; a task's value is the mean over the five substitutes and then the median over its ten frames.

\noindent\textbf{Masks.}
The cuts are diagnostic: the cut model generates actions with its own velocity.
Under the direct cut, the action tokens' positions still follow the instruction's length; a difference of one token changes \piH's action chunk on the motion dimensions by a median of 0.00684 in normalized units, never by a tenth of its largest value (\cref{supp:B0chk}).
PSG's guidance masks deny the main camera only the name tokens $N$ on the grounding pathway, and the action tokens the whole instruction $\ell$ on the direct pathway (\cref{supp:cut-results}); they only compute $v_g$ and $v_d$ at the same sampling state, and the gripper dimension keeps the original velocity.
The rows of \cref{supp:C4p2} that remove one effect change only the velocity of the six motion dimensions in PSG's sampler ($a=-1$, $b=0$ removes the grounding effect, $a=0$, $b=1$ the direct effect), and the gripper dimension keeps $v$.

\noindent\textbf{Cost.}
PSG adds one prefix pass per action chunk and two action-expert passes per Euler step.
A query takes 0.245 to 0.257\,s with PSG, against 0.154 to 0.172\,s for the released policy and 0.237\,s with CAG, on one 24\,GB card (\cref{supp:D5}).

\noindent\textbf{The Object grid.}
The Object grid places ten LIBERO-Object objects at their training places and at six new places each, with 10 episodes per object and place.
Start states in which the settled object touched more than the table or floor, or lay over 3\,cm off its place, were left out (four objects have 8 or 5 episodes at the grid cell x5y8).
The main paper's Fig.~7(a) shows each object at its 17\,cm or 27\,cm place; the direction of $v_g$ is read from the horizontal part of what $v_g$ adds to the executed actions, as its median cosine with the gripper-to-target direction over the episodes not yet closed.

\noindent\textbf{The push along the true direction.}
In \cref{supp:psg-source}, PSG's grounding effect is replaced by a push of 1.164 executed-action units toward the requested object's true position (times $d/5$\,cm at a distance $d<5$\,cm from the object), added to every action of the chunk.
On the 100 episodes of the main paper's Tab.~2 with a LIBERO-Goal target moved more than 30\,cm, Reach is 24\%, 50\% and 81\% and success 6\%, 13\% and 18\% for the released \piH, PSG and the push.

\noindent\textbf{Name tokens.}
$N$ denotes the tokens of every noun phrase of the instruction, found by a fixed rule that reads only the instruction: a phrase starts after ``the'', or after ``both'' or ``from'' before a content word, and runs to the next function word; color words join through ``and'', and a positional word before ``of'' is dropped.
An instruction with two objects to grasp puts both in $N$.
[D0: text to come.]}\par}
\suppcolfloat{B0chk}
\suppcolfloat{Gkv}
\subsection{Tables and Settings}
\label{supp:protocol-tables}
\input{sec/suppl/D4/algorithm}
PSG's settings in the main paper's comparisons are $K=10$, $a=5$ and $b=1$ (\cref{supp:D3}).
\piH\ predicts a chunk of 10 actions and executes the first five before the next replan.
With $a=b=0$, lines~9--10 return $v$, and Algorithm~\ref{alg:psg} is the original sampler.

[D4: text to come.]}\par}
\partfloats
\begin{supppage}
\setlength{\dblfloatsep}{0pt plus 2pt}
\suppfull{D3runs}
\suppfull{D2}
\end{supppage}
\begin{supppage}
\setlength{\dblfloatsep}{6pt plus 1pt}
\suppfull{D3}
\end{supppage}
\begin{supppage}
\setlength{\dblfloatsep}{6pt plus 1pt}
\suppfull{D6}
\suppw{D5}{0.485\textwidth}
\end{supppage}
\begin{table*}[tp]
\centering
\setlength\heavyrulewidth{0.25ex}
\captionsetup{position=top,skip=4pt}
\caption{\textbf{LIBERO per task.} Success of the four rows run here, 50 episodes per task; + CAG and + PSG on \piH. The main paper's Tab.~1 gives SmolVLA's LIBERO average as reported by its authors; this column runs it with 10 actions executed per replan (\cref{supp:D3}).
}
\label{supp:D1}\label{supp:D1a}
\begin{minipage}[t]{0.49\textwidth}\vspace{0pt}\centering\footnotesize
\setlength{\tabcolsep}{2.4pt}
\begin{tabular}{lrrr>{\columncolor[HTML]{EDEDED}}r}
\toprule
Task & SmolVLA & \piH & +\,CAG & +\,PSG\\
\midrule
\multicolumn{5}{l}{\textit{LIBERO-Spatial}}\\
\parbox[t]{4.45cm}{\raggedright 0 pick up the black bowl between the plate and the ramekin and place it on the plate} & 96 & 100 & 100 & 100\\
\parbox[t]{4.45cm}{\raggedright 1 pick up the black bowl next to the ramekin and place it on the plate} & 90 & 100 & 92 & 96\\
\parbox[t]{4.45cm}{\raggedright 2 pick up the black bowl from table center and place it on the plate} & 94 & 100 & 100 & 98\\
\parbox[t]{4.45cm}{\raggedright 3 pick up the black bowl on the cookie box and place it on the plate} & 88 & 98 & 98 & 100\\
\parbox[t]{4.45cm}{\raggedright 4 pick up the black bowl in the top drawer of the wooden cabinet and place it on the plate} & 66 & 96 & 88 & 84\\
\parbox[t]{4.45cm}{\raggedright 5 pick up the black bowl on the ramekin and place it on the plate} & 82 & 100 & 90 & 100\\
\parbox[t]{4.45cm}{\raggedright 6 pick up the black bowl next to the cookie box and place it on the plate} & 90 & 100 & 100 & 100\\
\parbox[t]{4.45cm}{\raggedright 7 pick up the black bowl on the stove and place it on the plate} & 82 & 100 & 90 & 94\\
\parbox[t]{4.45cm}{\raggedright 8 pick up the black bowl next to the plate and place it on the plate} & 86 & 100 & 94 & 98\\
\parbox[t]{4.45cm}{\raggedright 9 pick up the black bowl on the wooden cabinet and place it on the plate} & 76 & 98 & 94 & 96\\
\parbox[t]{4.45cm}{\raggedright \textbf{all tasks}} & 85.0 & 99.2 & 94.6 & 96.6\\
\midrule
\multicolumn{5}{l}{\textit{LIBERO-Object}}\\
\parbox[t]{4.45cm}{\raggedright 0 pick up the alphabet soup and place it in the basket} & 82 & 98 & 94 & 94\\
\parbox[t]{4.45cm}{\raggedright 1 pick up the cream cheese and place it in the basket} & 96 & 96 & 100 & 98\\
\parbox[t]{4.45cm}{\raggedright 2 pick up the salad dressing and place it in the basket} & 92 & 98 & 94 & 100\\
\parbox[t]{4.45cm}{\raggedright 3 pick up the bbq sauce and place it in the basket} & 96 & 98 & 94 & 92\\
\parbox[t]{4.45cm}{\raggedright 4 pick up the ketchup and place it in the basket} & 96 & 100 & 80 & 92\\
\parbox[t]{4.45cm}{\raggedright 5 pick up the tomato sauce and place it in the basket} & 76 & 94 & 96 & 94\\
\parbox[t]{4.45cm}{\raggedright 6 pick up the butter and place it in the basket} & 98 & 98 & 94 & 100\\
\parbox[t]{4.45cm}{\raggedright 7 pick up the milk and place it in the basket} & 96 & 100 & 96 & 100\\
\parbox[t]{4.45cm}{\raggedright 8 pick up the chocolate pudding and place it in the basket} & 80 & 100 & 94 & 98\\
\parbox[t]{4.45cm}{\raggedright 9 pick up the orange juice and place it in the basket} & 96 & 98 & 92 & 98\\
\parbox[t]{4.45cm}{\raggedright \textbf{all tasks}} & 90.8 & 98.0 & 93.4 & 96.6\\
\bottomrule
\end{tabular}
\end{minipage}\hfill
\begin{minipage}[t]{0.49\textwidth}\vspace{0pt}\centering\footnotesize
\setlength{\tabcolsep}{2.4pt}
\begin{tabular}{lrrr>{\columncolor[HTML]{EDEDED}}r}
\toprule
Task & SmolVLA & \piH & +\,CAG & +\,PSG\\
\midrule
\multicolumn{5}{l}{\textit{LIBERO-Goal}}\\
\parbox[t]{4.45cm}{\raggedright 0 open the middle drawer of the cabinet} & 98 & 96 & 88 & 98\\
\parbox[t]{4.45cm}{\raggedright 1 put the bowl on the stove} & 96 & 100 & 100 & 100\\
\parbox[t]{4.45cm}{\raggedright 2 put the wine bottle on top of the cabinet} & 94 & 98 & 88 & 100\\
\parbox[t]{4.45cm}{\raggedright 3 open the top drawer and put the bowl inside} & 80 & 94 & 78 & 84\\
\parbox[t]{4.45cm}{\raggedright 4 put the bowl on top of the cabinet} & 78 & 92 & 100 & 100\\
\parbox[t]{4.45cm}{\raggedright 5 push the plate to the front of the stove} & 96 & 100 & 94 & 100\\
\parbox[t]{4.45cm}{\raggedright 6 put the cream cheese in the bowl} & 56 & 98 & 90 & 98\\
\parbox[t]{4.45cm}{\raggedright 7 turn on the stove} & 100 & 100 & 84 & 94\\
\parbox[t]{4.45cm}{\raggedright 8 put the bowl on the plate} & 82 & 98 & 98 & 100\\
\parbox[t]{4.45cm}{\raggedright 9 put the wine bottle on the rack} & 78 & 94 & 78 & 100\\
\parbox[t]{4.45cm}{\raggedright \textbf{all tasks}} & 85.8 & 97.0 & 89.8 & 97.4\\
\midrule
\multicolumn{5}{l}{\textit{LIBERO-Long}}\\
\parbox[t]{4.45cm}{\raggedright 0 put both the alphabet soup and the tomato sauce in the basket} & 40 & 92 & 86 & 90\\
\parbox[t]{4.45cm}{\raggedright 1 put both the cream cheese box and the butter in the basket} & 94 & 98 & 92 & 100\\
\parbox[t]{4.45cm}{\raggedright 2 turn on the stove and put the moka pot on it} & 98 & 98 & 82 & 86\\
\parbox[t]{4.45cm}{\raggedright 3 put the black bowl in the bottom drawer of the cabinet and close it} & 100 & 96 & 72 & 92\\
\parbox[t]{4.45cm}{\raggedright 4 put the white mug on the left plate and put the yellow and white mug on the right plate} & 60 & 96 & 84 & 98\\
\parbox[t]{4.45cm}{\raggedright 5 pick up the book and place it in the back compartment of the caddy} & 82 & 100 & 78 & 100\\
\parbox[t]{4.45cm}{\raggedright 6 put the white mug on the plate and put the chocolate pudding to the right of the plate} & 54 & 84 & 64 & 88\\
\parbox[t]{4.45cm}{\raggedright 7 put both the alphabet soup and the cream cheese box in the basket} & 76 & 96 & 84 & 98\\
\parbox[t]{4.45cm}{\raggedright 8 put both moka pots on the stove} & 48 & 70 & 8 & 62\\
\parbox[t]{4.45cm}{\raggedright 9 put the yellow and white mug in the microwave and close it} & 82 & 94 & 80 & 86\\
\parbox[t]{4.45cm}{\raggedright \textbf{all tasks}} & 73.4 & 92.4 & 73.0 & 90.0\\
\bottomrule
\end{tabular}
\end{minipage}
\end{table*}

\begin{table*}[tp]
\centering
\setlength\heavyrulewidth{0.25ex}
\captionsetup{position=top,skip=4pt}
\caption{\textbf{LIBERO-PRO Position per task.} Reach (\textit{R}) and success (\textit{S}) of every row of the main paper's Tab.~1, 50 episodes per task; \textbf{all tasks}: that table's column; gray: requires training ($\dagger$); OFT: OpenVLA-OFT.
\ddag: the requested object stands on the other training place, so only the looser criterion counts its reaches (\textbf{conservative}: the Object tasks without \ddag); in parentheses, how far the object is moved; Goal task 0 has no grasp, so its Reach is a first contact with the middle drawer.
}
\label{supp:D1b}
\footnotesize
\setlength{\tabcolsep}{2.4pt}
\begin{tabular}{lr@{\hspace{1.5pt}}r@{\hspace{6pt}}r@{\hspace{1.5pt}}r@{\hspace{6pt}}r@{\hspace{1.5pt}}r@{\hspace{6pt}}r@{\hspace{1.5pt}}r@{\hspace{6pt}}r@{\hspace{1.5pt}}r@{\hspace{6pt}}>{\color{black!55}}r@{\hspace{1.5pt}}>{\color{black!55}}r@{\hspace{6pt}}>{\color{black!55}}r@{\hspace{1.5pt}}>{\color{black!55}}r@{\hspace{6pt}}r@{\hspace{1.5pt}}r@{\hspace{6pt}}r@{\hspace{1.5pt}}r@{\hspace{6pt}}>{\columncolor[HTML]{EDEDED}[3pt][0.75pt]}r@{\hspace{1.5pt}}>{\columncolor[HTML]{EDEDED}[0.75pt][3pt]}r}
\toprule
Task & \multicolumn{2}{c}{OFT} & \multicolumn{2}{c}{SmolVLA} & \multicolumn{2}{c}{TurboVLA} & \multicolumn{2}{c}{\piO-FAST} & \multicolumn{2}{c}{\piO} & \multicolumn{2}{c}{RSS$^\dagger$} & \multicolumn{2}{c}{APT$^\dagger$} & \multicolumn{2}{c}{\piH} & \multicolumn{2}{c}{+\,CAG} & \multicolumn{2}{c}{+\,PSG}\\
\cmidrule(lr){2-3}\cmidrule(lr){4-5}\cmidrule(lr){6-7}\cmidrule(lr){8-9}\cmidrule(lr){10-11}\cmidrule(lr){12-13}\cmidrule(lr){14-15}\cmidrule(lr){16-17}\cmidrule(lr){18-19}\cmidrule(lr){20-21}
 & {\scriptsize\itshape R} & {\scriptsize\itshape S} & {\scriptsize\itshape R} & {\scriptsize\itshape S} & {\scriptsize\itshape R} & {\scriptsize\itshape S} & {\scriptsize\itshape R} & {\scriptsize\itshape S} & {\scriptsize\itshape R} & {\scriptsize\itshape S} & {\scriptsize\itshape R} & {\scriptsize\itshape S} & {\scriptsize\itshape R} & {\scriptsize\itshape S} & {\scriptsize\itshape R} & {\scriptsize\itshape S} & {\scriptsize\itshape R} & {\scriptsize\itshape S} & {\scriptsize\itshape R} & {\scriptsize\itshape S}\\
\midrule
\multicolumn{21}{l}{\textit{LIBERO-PRO Position Object}}\\
\parbox[t]{3.95cm}{\raggedright 0 pick up the alphabet soup and place it in the basket (17.9\,cm)} & 0 & 0 & 0 & 0 & 0 & 0 & 0 & 0 & 0 & 0 & 98 & 26 & 0 & 0 & 42 & 38 & 10 & 10 & 84 & 72\\
\parbox[t]{3.95cm}{\raggedright 1 pick up the cream cheese and place it in the basket\textsuperscript{\ddag} (22.0\,cm)} & 2 & 4 & 0 & 0 & 0 & 0 & 0 & 0 & 0 & 0 & 100 & 100 & 100 & 100 & 100 & 94 & 100 & 98 & 100 & 100\\
\parbox[t]{3.95cm}{\raggedright 2 pick up the salad dressing and place it in the basket (25.1\,cm)} & 0 & 0 & 0 & 0 & 0 & 0 & 0 & 0 & 0 & 0 & 98 & 50 & 0 & 0 & 0 & 0 & 0 & 0 & 0 & 0\\
\parbox[t]{3.95cm}{\raggedright 3 pick up the bbq sauce and place it in the basket\textsuperscript{\ddag} (22.0\,cm)} & 0 & 0 & 0 & 0 & 0 & 0 & 0 & 0 & 0 & 0 & 98 & 86 & 0 & 0 & 0 & 0 & 0 & 0 & 0 & 0\\
\parbox[t]{3.95cm}{\raggedright 4 pick up the ketchup and place it in the basket (38.2\,cm)} & 0 & 0 & 0 & 0 & 0 & 0 & 0 & 0 & 0 & 0 & 0 & 0 & 0 & 0 & 0 & 0 & 0 & 0 & 0 & 0\\
\parbox[t]{3.95cm}{\raggedright 5 pick up the tomato sauce and place it in the basket\textsuperscript{\ddag} (22.0\,cm)} & 0 & 2 & 0 & 0 & 8 & 2 & 0 & 0 & 0 & 0 & 100 & 92 & 0 & 0 & 0 & 0 & 0 & 0 & 2 & 0\\
\parbox[t]{3.95cm}{\raggedright 6 pick up the butter and place it in the basket\textsuperscript{\ddag} (22.0\,cm)} & 0 & 0 & 0 & 0 & 0 & 0 & 0 & 0 & 0 & 2 & 100 & 100 & 4 & 2 & 50 & 28 & 62 & 36 & 100 & 100\\
\parbox[t]{3.95cm}{\raggedright 7 pick up the milk and place it in the basket\textsuperscript{\ddag} (22.0\,cm)} & 0 & 0 & 0 & 0 & 0 & 0 & 0 & 0 & 0 & 0 & 100 & 92 & 4 & 0 & 0 & 0 & 0 & 0 & 92 & 82\\
\parbox[t]{3.95cm}{\raggedright 8 pick up the chocolate pudding and place it in the basket (22.4\,cm)} & 0 & 0 & 0 & 0 & 0 & 0 & 0 & 0 & 0 & 0 & 0 & 0 & 16 & 4 & 0 & 0 & 0 & 0 & 48 & 40\\
\parbox[t]{3.95cm}{\raggedright 9 pick up the orange juice and place it in the basket (25.6\,cm)} & 0 & 0 & 0 & 0 & 0 & 0 & 0 & 0 & 0 & 0 & 94 & 2 & 0 & 0 & 28 & 6 & 94 & 20 & 100 & 14\\
\parbox[t]{3.95cm}{\raggedright \textbf{all tasks}} & 0.2 & 0.6 & 0.0 & 0.0 & 0.8 & 0.2 & 0.0 & 0.0 & 0.0 & 0.2 & 78.8 & 54.8 & 12.4 & 10.6 & 22.0 & 16.6 & 26.6 & 16.4 & 52.6 & 40.8\\
\parbox[t]{3.95cm}{\raggedright \textbf{all tasks, conservative criterion}} & 0.0 &  & 0.0 &  & 0.0 &  & 0.0 &  & 0.0 &  & 58.0 &  & 3.2 &  & 14.0 &  & 20.8 &  & 46.4 & \\
\bottomrule
\end{tabular}

\par\medskip
\footnotesize
\setlength{\tabcolsep}{2.4pt}
\begin{tabular}{lr@{\hspace{1.5pt}}r@{\hspace{6pt}}r@{\hspace{1.5pt}}r@{\hspace{6pt}}r@{\hspace{1.5pt}}r@{\hspace{6pt}}r@{\hspace{1.5pt}}r@{\hspace{6pt}}r@{\hspace{1.5pt}}r@{\hspace{6pt}}>{\color{black!55}}r@{\hspace{1.5pt}}>{\color{black!55}}r@{\hspace{6pt}}>{\color{black!55}}r@{\hspace{1.5pt}}>{\color{black!55}}r@{\hspace{6pt}}r@{\hspace{1.5pt}}r@{\hspace{6pt}}r@{\hspace{1.5pt}}r@{\hspace{6pt}}>{\columncolor[HTML]{EDEDED}[3pt][0.75pt]}r@{\hspace{1.5pt}}>{\columncolor[HTML]{EDEDED}[0.75pt][3pt]}r}
\toprule
Task & \multicolumn{2}{c}{OFT} & \multicolumn{2}{c}{SmolVLA} & \multicolumn{2}{c}{TurboVLA} & \multicolumn{2}{c}{\piO-FAST} & \multicolumn{2}{c}{\piO} & \multicolumn{2}{c}{RSS$^\dagger$} & \multicolumn{2}{c}{APT$^\dagger$} & \multicolumn{2}{c}{\piH} & \multicolumn{2}{c}{+\,CAG} & \multicolumn{2}{c}{+\,PSG}\\
\cmidrule(lr){2-3}\cmidrule(lr){4-5}\cmidrule(lr){6-7}\cmidrule(lr){8-9}\cmidrule(lr){10-11}\cmidrule(lr){12-13}\cmidrule(lr){14-15}\cmidrule(lr){16-17}\cmidrule(lr){18-19}\cmidrule(lr){20-21}
 & {\scriptsize\itshape R} & {\scriptsize\itshape S} & {\scriptsize\itshape R} & {\scriptsize\itshape S} & {\scriptsize\itshape R} & {\scriptsize\itshape S} & {\scriptsize\itshape R} & {\scriptsize\itshape S} & {\scriptsize\itshape R} & {\scriptsize\itshape S} & {\scriptsize\itshape R} & {\scriptsize\itshape S} & {\scriptsize\itshape R} & {\scriptsize\itshape S} & {\scriptsize\itshape R} & {\scriptsize\itshape S} & {\scriptsize\itshape R} & {\scriptsize\itshape S} & {\scriptsize\itshape R} & {\scriptsize\itshape S}\\
\midrule
\multicolumn{21}{l}{\textit{LIBERO-PRO Position Goal}}\\
\parbox[t]{3.95cm}{\raggedright 0 open the middle drawer of the cabinet} & 0 & 0 & 0 & 0 & 0 & 0 & 0 & 0 & 0 & 0 & 0 & 0 & 0 & 0 & 0 & 0 & 0 & 0 & 0 & 0\\
\parbox[t]{3.95cm}{\raggedright 1 put the bowl on the stove} & 98 & 0 & 0 & 0 & 0 & 0 & 84 & 0 & 6 & 0 & 98 & 80 & 100 & 12 & 100 & 84 & 100 & 54 & 100 & 40\\
\parbox[t]{3.95cm}{\raggedright 2 put the wine bottle on top of the cabinet} & 0 & 0 & 0 & 0 & 18 & 0 & 58 & 0 & 0 & 0 & 100 & 0 & 90 & 0 & 100 & 0 & 100 & 0 & 100 & 0\\
\parbox[t]{3.95cm}{\raggedright 3 open the top drawer and put the bowl inside} & 100 & 0 & 100 & 0 & 100 & 10 & 100 & 26 & 94 & 50 & 82 & 74 & 100 & 52 & 100 & 98 & 100 & 84 & 100 & 92\\
\parbox[t]{3.95cm}{\raggedright 4 put the bowl on top of the cabinet} & 40 & 0 & 54 & 0 & 76 & 0 & 100 & 0 & 38 & 0 & 100 & 0 & 100 & 0 & 100 & 0 & 100 & 0 & 100 & 0\\
\parbox[t]{3.95cm}{\raggedright 5 push the plate to the front of the stove} & 0 & 0 & 0 & 0 & 0 & 0 & 4 & 0 & 0 & 0 & 22 & 0 & 2 & 0 & 60 & 0 & 62 & 0 & 98 & 0\\
\parbox[t]{3.95cm}{\raggedright 6 put the cream cheese in the bowl} & 0 & 0 & 0 & 0 & 0 & 0 & 0 & 0 & 0 & 0 & 58 & 8 & 80 & 4 & 0 & 0 & 0 & 0 & 80 & 42\\
\parbox[t]{3.95cm}{\raggedright 7 turn on the stove} & 0 & 0 & 0 & 0 & 0 & 0 & 0 & 4 & 0 & 90 & 100 & 88 & 6 & 4 & 98 & 84 & 74 & 66 & 94 & 100\\
\parbox[t]{3.95cm}{\raggedright 8 put the bowl on the plate} & 58 & 0 & 94 & 0 & 100 & 0 & 100 & 14 & 86 & 0 & 100 & 52 & 100 & 0 & 100 & 66 & 100 & 62 & 100 & 88\\
\parbox[t]{3.95cm}{\raggedright 9 put the wine bottle on the rack} & 0 & 0 & 0 & 0 & 22 & 0 & 40 & 2 & 4 & 0 & 8 & 0 & 62 & 0 & 100 & 0 & 32 & 0 & 98 & 46\\
\parbox[t]{3.95cm}{\raggedright \textbf{all tasks}} & 29.6 & 0.0 & 24.8 & 0.0 & 31.6 & 1.0 & 48.6 & 4.6 & 22.8 & 14.0 & 66.8 & 30.2 & 64.0 & 7.2 & 75.8 & 33.2 & 66.8 & 26.6 & 87.0 & 40.8\\
\bottomrule
\end{tabular}

\end{table*}
\begin{table*}[tp]
\centering
\setlength\heavyrulewidth{0.25ex}
\ContinuedFloat
\captionsetup{position=top,skip=4pt}
\caption[]{(continued)}
\footnotesize
\setlength{\tabcolsep}{2.4pt}
\begin{tabular}{lr@{\hspace{1.5pt}}r@{\hspace{6pt}}r@{\hspace{1.5pt}}r@{\hspace{6pt}}r@{\hspace{1.5pt}}r@{\hspace{6pt}}r@{\hspace{1.5pt}}r@{\hspace{6pt}}r@{\hspace{1.5pt}}r@{\hspace{6pt}}>{\color{black!55}}r@{\hspace{1.5pt}}>{\color{black!55}}r@{\hspace{6pt}}>{\color{black!55}}r@{\hspace{1.5pt}}>{\color{black!55}}r@{\hspace{6pt}}r@{\hspace{1.5pt}}r@{\hspace{6pt}}r@{\hspace{1.5pt}}r@{\hspace{6pt}}>{\columncolor[HTML]{EDEDED}[3pt][0.75pt]}r@{\hspace{1.5pt}}>{\columncolor[HTML]{EDEDED}[0.75pt][3pt]}r}
\toprule
Task & \multicolumn{2}{c}{OFT} & \multicolumn{2}{c}{SmolVLA} & \multicolumn{2}{c}{TurboVLA} & \multicolumn{2}{c}{\piO-FAST} & \multicolumn{2}{c}{\piO} & \multicolumn{2}{c}{RSS$^\dagger$} & \multicolumn{2}{c}{APT$^\dagger$} & \multicolumn{2}{c}{\piH} & \multicolumn{2}{c}{+\,CAG} & \multicolumn{2}{c}{+\,PSG}\\
\cmidrule(lr){2-3}\cmidrule(lr){4-5}\cmidrule(lr){6-7}\cmidrule(lr){8-9}\cmidrule(lr){10-11}\cmidrule(lr){12-13}\cmidrule(lr){14-15}\cmidrule(lr){16-17}\cmidrule(lr){18-19}\cmidrule(lr){20-21}
 & {\scriptsize\itshape R} & {\scriptsize\itshape S} & {\scriptsize\itshape R} & {\scriptsize\itshape S} & {\scriptsize\itshape R} & {\scriptsize\itshape S} & {\scriptsize\itshape R} & {\scriptsize\itshape S} & {\scriptsize\itshape R} & {\scriptsize\itshape S} & {\scriptsize\itshape R} & {\scriptsize\itshape S} & {\scriptsize\itshape R} & {\scriptsize\itshape S} & {\scriptsize\itshape R} & {\scriptsize\itshape S} & {\scriptsize\itshape R} & {\scriptsize\itshape S} & {\scriptsize\itshape R} & {\scriptsize\itshape S}\\
\midrule
\multicolumn{21}{l}{\textit{LIBERO-PRO Position Spatial}}\\
\parbox[t]{3.95cm}{\raggedright 0 pick up the black bowl between the plate and the ramekin and place it on the plate} & 100 & 94 & 100 & 0 & 100 & 28 & 100 & 26 & 100 & 0 & 100 & 96 & 100 & 46 & 100 & 92 & 100 & 98 & 100 & 98\\
\parbox[t]{3.95cm}{\raggedright 1 pick up the black bowl next to the ramekin and place it on the plate} & 56 & 0 & 100 & 0 & 16 & 0 & 100 & 0 & 100 & 0 & 100 & 60 & 100 & 44 & 100 & 22 & 100 & 50 & 98 & 82\\
\parbox[t]{3.95cm}{\raggedright 2 pick up the black bowl from table center and place it on the plate} & 100 & 38 & 100 & 0 & 100 & 50 & 100 & 98 & 100 & 0 & 100 & 100 & 100 & 96 & 100 & 100 & 100 & 100 & 100 & 100\\
\parbox[t]{3.95cm}{\raggedright 3 pick up the black bowl on the cookie box and place it on the plate} & 0 & 0 & 4 & 0 & 0 & 0 & 0 & 0 & 0 & 0 & 4 & 2 & 4 & 0 & 0 & 0 & 0 & 0 & 2 & 0\\
\parbox[t]{3.95cm}{\raggedright 4 pick up the black bowl in the top drawer of the wooden cabinet and place it on the plate} & 100 & 28 & 100 & 0 & 100 & 0 & 100 & 50 & 100 & 0 & 100 & 82 & 100 & 16 & 100 & 54 & 100 & 38 & 100 & 38\\
\parbox[t]{3.95cm}{\raggedright 5 pick up the black bowl on the ramekin and place it on the plate} & 100 & 60 & 100 & 0 & 82 & 14 & 100 & 16 & 100 & 2 & 100 & 100 & 100 & 64 & 100 & 86 & 100 & 78 & 100 & 96\\
\parbox[t]{3.95cm}{\raggedright 6 pick up the black bowl next to the cookie box and place it on the plate} & 100 & 0 & 84 & 0 & 14 & 0 & 100 & 0 & 100 & 0 & 100 & 0 & 100 & 0 & 100 & 0 & 100 & 0 & 100 & 0\\
\parbox[t]{3.95cm}{\raggedright 7 pick up the black bowl on the stove and place it on the plate} & 100 & 0 & 100 & 0 & 100 & 0 & 100 & 0 & 100 & 0 & 100 & 4 & 100 & 30 & 100 & 0 & 100 & 0 & 100 & 10\\
\parbox[t]{3.95cm}{\raggedright 8 pick up the black bowl next to the plate and place it on the plate} & 98 & 56 & 100 & 0 & 100 & 70 & 100 & 100 & 100 & 82 & 100 & 88 & 80 & 90 & 100 & 98 & 100 & 94 & 100 & 100\\
\parbox[t]{3.95cm}{\raggedright 9 pick up the black bowl on the wooden cabinet and place it on the plate} & 100 & 0 & 100 & 0 & 100 & 0 & 100 & 0 & 100 & 0 & 100 & 0 & 100 & 0 & 100 & 0 & 100 & 0 & 100 & 0\\
\parbox[t]{3.95cm}{\raggedright \textbf{all tasks}} & 85.4 & 27.6 & 88.8 & 0.0 & 71.2 & 16.2 & 90.0 & 29.0 & 90.0 & 8.4 & 90.4 & 53.2 & 88.4 & 38.6 & 90.0 & 45.2 & 90.0 & 45.8 & 90.0 & 52.4\\
\bottomrule
\end{tabular}

\end{table*}

\begin{table*}[tp]
\centering
\setlength\heavyrulewidth{0.25ex}
\captionsetup{position=top,skip=4pt}
\caption{\textbf{LIBERO-PRO Task per task.} Reach (\textit{R}) and success (\textit{S}) per suite, 50 episodes per task; the main paper's Tab.~1 prints the mean of the three suites. Gray: requires training ($\dagger$).
}
\label{supp:D1c}
\footnotesize
\setlength{\tabcolsep}{2.4pt}
\begin{tabular}{lr@{\hspace{1.5pt}}r@{\hspace{6pt}}r@{\hspace{1.5pt}}r@{\hspace{6pt}}r@{\hspace{1.5pt}}r@{\hspace{6pt}}r@{\hspace{1.5pt}}r@{\hspace{6pt}}r@{\hspace{1.5pt}}r@{\hspace{6pt}}>{\color{black!55}}r@{\hspace{1.5pt}}>{\color{black!55}}r@{\hspace{6pt}}>{\color{black!55}}r@{\hspace{1.5pt}}>{\color{black!55}}r@{\hspace{6pt}}r@{\hspace{1.5pt}}r@{\hspace{6pt}}r@{\hspace{1.5pt}}r@{\hspace{6pt}}>{\columncolor[HTML]{EDEDED}[3pt][0.75pt]}r@{\hspace{1.5pt}}>{\columncolor[HTML]{EDEDED}[0.75pt][3pt]}r}
\toprule
Task & \multicolumn{2}{c}{OFT} & \multicolumn{2}{c}{SmolVLA} & \multicolumn{2}{c}{TurboVLA} & \multicolumn{2}{c}{\piO-FAST} & \multicolumn{2}{c}{\piO} & \multicolumn{2}{c}{RSS$^\dagger$} & \multicolumn{2}{c}{APT$^\dagger$} & \multicolumn{2}{c}{\piH} & \multicolumn{2}{c}{+\,CAG} & \multicolumn{2}{c}{+\,PSG}\\
\cmidrule(lr){2-3}\cmidrule(lr){4-5}\cmidrule(lr){6-7}\cmidrule(lr){8-9}\cmidrule(lr){10-11}\cmidrule(lr){12-13}\cmidrule(lr){14-15}\cmidrule(lr){16-17}\cmidrule(lr){18-19}\cmidrule(lr){20-21}
 & {\scriptsize\itshape R} & {\scriptsize\itshape S} & {\scriptsize\itshape R} & {\scriptsize\itshape S} & {\scriptsize\itshape R} & {\scriptsize\itshape S} & {\scriptsize\itshape R} & {\scriptsize\itshape S} & {\scriptsize\itshape R} & {\scriptsize\itshape S} & {\scriptsize\itshape R} & {\scriptsize\itshape S} & {\scriptsize\itshape R} & {\scriptsize\itshape S} & {\scriptsize\itshape R} & {\scriptsize\itshape S} & {\scriptsize\itshape R} & {\scriptsize\itshape S} & {\scriptsize\itshape R} & {\scriptsize\itshape S}\\
\midrule
\multicolumn{21}{l}{\textit{LIBERO-PRO Task Object}}\\
\parbox[t]{3.95cm}{\raggedright 0 pick the cream cheese and place it in the basket} & 0 & 0 & 0 & 0 & 0 & 0 & 0 & 0 & 0 & 0 & 14 & 26 & 86 & 56 & 10 & 8 & 24 & 16 & 50 & 34\\
\parbox[t]{3.95cm}{\raggedright 1 pick the alphabet soup and place it in the basket} & 0 & 0 & 100 & 20 & 0 & 0 & 100 & 98 & 100 & 94 & 100 & 98 & 100 & 86 & 100 & 94 & 100 & 86 & 100 & 92\\
\parbox[t]{3.95cm}{\raggedright 2 pick the tomato sauce and place it in the basket} & 0 & 0 & 0 & 0 & 0 & 0 & 0 & 0 & 0 & 0 & 0 & 0 & 0 & 0 & 0 & 0 & 0 & 0 & 0 & 0\\
\parbox[t]{3.95cm}{\raggedright 3 pick the ketchup and place it in the basket} & 0 & 0 & 0 & 0 & 0 & 0 & 0 & 0 & 0 & 0 & 68 & 26 & 0 & 0 & 0 & 0 & 0 & 0 & 0 & 0\\
\parbox[t]{3.95cm}{\raggedright 4 pick the milk and place it in the basket} & 0 & 0 & 0 & 0 & 0 & 0 & 0 & 0 & 0 & 0 & 0 & 0 & 0 & 0 & 0 & 0 & 0 & 0 & 0 & 0\\
\parbox[t]{3.95cm}{\raggedright 5 pick the bbq sauce and place it in the basket} & 0 & 0 & 0 & 0 & 0 & 0 & 0 & 0 & 0 & 0 & 0 & 0 & 0 & 0 & 0 & 0 & 0 & 0 & 0 & 0\\
\parbox[t]{3.95cm}{\raggedright 6 pick the orange juice and place it in the basket} & 0 & 0 & 0 & 0 & 0 & 0 & 0 & 0 & 0 & 0 & 100 & 22 & 0 & 0 & 0 & 0 & 0 & 0 & 14 & 2\\
\parbox[t]{3.95cm}{\raggedright 7 pick the butter and place it in the basket} & 0 & 0 & 0 & 0 & 0 & 0 & 0 & 0 & 0 & 0 & 90 & 76 & 0 & 0 & 20 & 4 & 4 & 0 & 0 & 0\\
\parbox[t]{3.95cm}{\raggedright 8 pick the salad dressing and place it in the basket} & 0 & 0 & 0 & 0 & 0 & 0 & 0 & 0 & 0 & 0 & 58 & 16 & 0 & 0 & 0 & 0 & 10 & 4 & 66 & 16\\
\parbox[t]{3.95cm}{\raggedright 9 pick the chocolate pudding and place it in the basket} & 0 & 0 & 0 & 0 & 0 & 0 & 0 & 0 & 0 & 0 & 0 & 0 & 0 & 0 & 0 & 0 & 0 & 0 & 0 & 4\\
\parbox[t]{3.95cm}{\raggedright \textbf{all tasks}} & 0.0 & 0.0 & 10.0 & 2.0 & 0.0 & 0.0 & 10.0 & 9.8 & 10.0 & 9.4 & 43.0 & 26.4 & 18.6 & 14.2 & 13.0 & 10.6 & 13.8 & 10.6 & 23.0 & 14.8\\
\bottomrule
\end{tabular}

\end{table*}
\begin{table*}[tp]
\centering
\setlength\heavyrulewidth{0.25ex}
\ContinuedFloat
\captionsetup{position=top,skip=4pt}
\caption[]{(continued)}
\footnotesize
\setlength{\tabcolsep}{2.4pt}
\begin{tabular}{lr@{\hspace{1.5pt}}r@{\hspace{6pt}}r@{\hspace{1.5pt}}r@{\hspace{6pt}}r@{\hspace{1.5pt}}r@{\hspace{6pt}}r@{\hspace{1.5pt}}r@{\hspace{6pt}}r@{\hspace{1.5pt}}r@{\hspace{6pt}}>{\color{black!55}}r@{\hspace{1.5pt}}>{\color{black!55}}r@{\hspace{6pt}}>{\color{black!55}}r@{\hspace{1.5pt}}>{\color{black!55}}r@{\hspace{6pt}}r@{\hspace{1.5pt}}r@{\hspace{6pt}}r@{\hspace{1.5pt}}r@{\hspace{6pt}}>{\columncolor[HTML]{EDEDED}[3pt][0.75pt]}r@{\hspace{1.5pt}}>{\columncolor[HTML]{EDEDED}[0.75pt][3pt]}r}
\toprule
Task & \multicolumn{2}{c}{OFT} & \multicolumn{2}{c}{SmolVLA} & \multicolumn{2}{c}{TurboVLA} & \multicolumn{2}{c}{\piO-FAST} & \multicolumn{2}{c}{\piO} & \multicolumn{2}{c}{RSS$^\dagger$} & \multicolumn{2}{c}{APT$^\dagger$} & \multicolumn{2}{c}{\piH} & \multicolumn{2}{c}{+\,CAG} & \multicolumn{2}{c}{+\,PSG}\\
\cmidrule(lr){2-3}\cmidrule(lr){4-5}\cmidrule(lr){6-7}\cmidrule(lr){8-9}\cmidrule(lr){10-11}\cmidrule(lr){12-13}\cmidrule(lr){14-15}\cmidrule(lr){16-17}\cmidrule(lr){18-19}\cmidrule(lr){20-21}
 & {\scriptsize\itshape R} & {\scriptsize\itshape S} & {\scriptsize\itshape R} & {\scriptsize\itshape S} & {\scriptsize\itshape R} & {\scriptsize\itshape S} & {\scriptsize\itshape R} & {\scriptsize\itshape S} & {\scriptsize\itshape R} & {\scriptsize\itshape S} & {\scriptsize\itshape R} & {\scriptsize\itshape S} & {\scriptsize\itshape R} & {\scriptsize\itshape S} & {\scriptsize\itshape R} & {\scriptsize\itshape S} & {\scriptsize\itshape R} & {\scriptsize\itshape S} & {\scriptsize\itshape R} & {\scriptsize\itshape S}\\
\midrule
\multicolumn{21}{l}{\textit{LIBERO-PRO Task Goal}}\\
\parbox[t]{3.95cm}{\raggedright 0 open the bottom drawer of the cabinet} & 0 & 0 & 0 & 2 & 0 & 0 & 0 & 0 & 2 & 2 & 0 & 0 & 0 & 0 & 0 & 2 & 0 & 0 & 0 & 0\\
\parbox[t]{3.95cm}{\raggedright 1 put the plate on the stove} & 12 & 0 & 0 & 0 & 0 & 0 & 38 & 0 & 0 & 0 & 20 & 0 & 60 & 0 & 26 & 10 & 24 & 8 & 94 & 40\\
\parbox[t]{3.95cm}{\raggedright 2 put the wine bottle in the bowl} & 70 & 0 & 0 & 0 & 2 & 0 & 42 & 2 & 20 & 0 & 100 & 0 & 100 & 10 & 100 & 32 & 100 & 38 & 100 & 46\\
\parbox[t]{3.95cm}{\raggedright 3 open the top layer of the drawer and put the cream cheese inside} & 100 & 0 & 100 & 0 & 0 & 0 & 100 & 0 & 88 & 0 & 90 & 78 & 76 & 12 & 98 & 0 & 90 & 8 & 90 & 22\\
\parbox[t]{3.95cm}{\raggedright 4 put the plate on the top of the drawer} & 64 & 0 & 0 & 0 & 0 & 0 & 10 & 0 & 0 & 0 & 32 & 0 & 28 & 0 & 18 & 10 & 22 & 14 & 40 & 18\\
\parbox[t]{3.95cm}{\raggedright 5 push the cream cheese to the front of the stove} & 4 & 2 & 0 & 6 & 22 & 0 & 24 & 0 & 0 & 12 & 0 & 0 & 24 & 30 & 18 & 2 & 16 & 0 & 8 & 0\\
\parbox[t]{3.95cm}{\raggedright 6 put the wine bottle in the bowl} & 70 & 0 & 0 & 0 & 0 & 0 & 48 & 0 & 30 & 0 & 100 & 0 & 100 & 12 & 100 & 22 & 100 & 30 & 100 & 42\\
\parbox[t]{3.95cm}{\raggedright 7 turn off the stove} & 98 & 92 & 98 & 90 & 92 & 90 & 96 & 94 & 94 & 82 & 98 & 96 & 98 & 86 & 94 & 94 & 98 & 84 & 84 & 76\\
\parbox[t]{3.95cm}{\raggedright 8 put the wine bottle on the plate} & 34 & 0 & 12 & 0 & 100 & 0 & 84 & 0 & 98 & 0 & 100 & 0 & 100 & 14 & 100 & 62 & 100 & 42 & 100 & 86\\
\parbox[t]{3.95cm}{\raggedright 9 put the cream cheese on the rack} & 0 & 0 & 16 & 0 & 98 & 0 & 100 & 0 & 0 & 0 & 100 & 0 & 98 & 0 & 100 & 0 & 100 & 0 & 100 & 0\\
\parbox[t]{3.95cm}{\raggedright \textbf{all tasks}} & 45.2 & 9.4 & 22.6 & 9.8 & 31.4 & 9.0 & 54.2 & 9.6 & 33.2 & 9.6 & 64.0 & 17.4 & 68.4 & 16.4 & 65.4 & 23.4 & 65.0 & 22.4 & 71.6 & 33.0\\
\bottomrule
\end{tabular}

\par\medskip
\footnotesize
\setlength{\tabcolsep}{2.4pt}
\begin{tabular}{lr@{\hspace{1.5pt}}r@{\hspace{6pt}}r@{\hspace{1.5pt}}r@{\hspace{6pt}}r@{\hspace{1.5pt}}r@{\hspace{6pt}}r@{\hspace{1.5pt}}r@{\hspace{6pt}}r@{\hspace{1.5pt}}r@{\hspace{6pt}}>{\color{black!55}}r@{\hspace{1.5pt}}>{\color{black!55}}r@{\hspace{6pt}}>{\color{black!55}}r@{\hspace{1.5pt}}>{\color{black!55}}r@{\hspace{6pt}}r@{\hspace{1.5pt}}r@{\hspace{6pt}}r@{\hspace{1.5pt}}r@{\hspace{6pt}}>{\columncolor[HTML]{EDEDED}[3pt][0.75pt]}r@{\hspace{1.5pt}}>{\columncolor[HTML]{EDEDED}[0.75pt][3pt]}r}
\toprule
Task & \multicolumn{2}{c}{OFT} & \multicolumn{2}{c}{SmolVLA} & \multicolumn{2}{c}{TurboVLA} & \multicolumn{2}{c}{\piO-FAST} & \multicolumn{2}{c}{\piO} & \multicolumn{2}{c}{RSS$^\dagger$} & \multicolumn{2}{c}{APT$^\dagger$} & \multicolumn{2}{c}{\piH} & \multicolumn{2}{c}{+\,CAG} & \multicolumn{2}{c}{+\,PSG}\\
\cmidrule(lr){2-3}\cmidrule(lr){4-5}\cmidrule(lr){6-7}\cmidrule(lr){8-9}\cmidrule(lr){10-11}\cmidrule(lr){12-13}\cmidrule(lr){14-15}\cmidrule(lr){16-17}\cmidrule(lr){18-19}\cmidrule(lr){20-21}
 & {\scriptsize\itshape R} & {\scriptsize\itshape S} & {\scriptsize\itshape R} & {\scriptsize\itshape S} & {\scriptsize\itshape R} & {\scriptsize\itshape S} & {\scriptsize\itshape R} & {\scriptsize\itshape S} & {\scriptsize\itshape R} & {\scriptsize\itshape S} & {\scriptsize\itshape R} & {\scriptsize\itshape S} & {\scriptsize\itshape R} & {\scriptsize\itshape S} & {\scriptsize\itshape R} & {\scriptsize\itshape S} & {\scriptsize\itshape R} & {\scriptsize\itshape S} & {\scriptsize\itshape R} & {\scriptsize\itshape S}\\
\midrule
\multicolumn{21}{l}{\textit{LIBERO-PRO Task Spatial}}\\
\parbox[t]{3.95cm}{\raggedright 0 pick the akita black bowl not between the plate and the ramekin and place it on the plate} & 0 & 0 & 2 & 0 & 0 & 0 & 0 & 0 & 0 & 0 & 6 & 2 & 0 & 0 & 0 & 2 & 0 & 0 & 2 & 2\\
\parbox[t]{3.95cm}{\raggedright 1 pick the akita black bowl next to the cookie box and place it on the plate} & 0 & 0 & 0 & 0 & 0 & 0 & 100 & 100 & 100 & 100 & 100 & 100 & 100 & 100 & 100 & 98 & 100 & 98 & 98 & 96\\
\parbox[t]{3.95cm}{\raggedright 2 pick the akita black bowl next to the plate and place it on the plate} & 0 & 0 & 0 & 0 & 0 & 2 & 0 & 0 & 16 & 16 & 0 & 0 & 48 & 50 & 0 & 2 & 0 & 0 & 0 & 0\\
\parbox[t]{3.95cm}{\raggedright 3 pick the akita black bowl on the top of the cabinet and place it on the plate} & 0 & 0 & 0 & 0 & 0 & 0 & 100 & 98 & 0 & 0 & 88 & 86 & 0 & 0 & 18 & 18 & 46 & 46 & 52 & 52\\
\parbox[t]{3.95cm}{\raggedright 4 pick the akita black bowl on the top of the wooden cabinet and place it on the plate} & 0 & 0 & 0 & 0 & 0 & 0 & 2 & 2 & 8 & 6 & 0 & 0 & 0 & 0 & 0 & 0 & 0 & 0 & 6 & 6\\
\parbox[t]{3.95cm}{\raggedright 5 pick the akita black bowl on the cookie box and place it on the plate} & 2 & 0 & 100 & 58 & 18 & 10 & 100 & 100 & 100 & 100 & 100 & 98 & 100 & 96 & 100 & 98 & 100 & 98 & 100 & 98\\
\parbox[t]{3.95cm}{\raggedright 6 pick the akita black bowl on the stove and place it on the plate} & 2 & 0 & 0 & 0 & 0 & 0 & 100 & 94 & 100 & 98 & 100 & 98 & 100 & 98 & 92 & 92 & 100 & 100 & 92 & 90\\
\parbox[t]{3.95cm}{\raggedright 7 pick the akita black bowl on the top of the cabinet and place it on the plate} & 100 & 96 & 0 & 0 & 94 & 76 & 100 & 98 & 88 & 86 & 100 & 98 & 100 & 96 & 98 & 98 & 98 & 96 & 90 & 86\\
\parbox[t]{3.95cm}{\raggedright 8 pick the akita black bowl next to the ramekin and place it on the plate} & 0 & 0 & 4 & 0 & 0 & 0 & 34 & 34 & 46 & 44 & 100 & 94 & 86 & 80 & 46 & 40 & 98 & 96 & 26 & 26\\
\parbox[t]{3.95cm}{\raggedright 9 pick the akita black bowl on the stove and place it on the plate} & 100 & 100 & 2 & 0 & 94 & 84 & 100 & 98 & 100 & 100 & 100 & 98 & 100 & 100 & 100 & 100 & 100 & 100 & 100 & 94\\
\parbox[t]{3.95cm}{\raggedright \textbf{all tasks}} & 20.4 & 19.6 & 10.8 & 5.8 & 20.6 & 17.2 & 63.6 & 62.4 & 55.8 & 55.0 & 69.4 & 67.4 & 63.4 & 62.0 & 55.4 & 54.8 & 64.2 & 63.4 & 56.6 & 55.0\\
\bottomrule
\end{tabular}

\end{table*}
{\small D1: not in place yet.}}\par}
\begin{table*}[tp]
\footnotesize
\centering
\setlength\heavyrulewidth{0.25ex}
\setlength{\tabcolsep}{2.2pt}
\captionsetup{position=top,skip=4pt}
\caption{\textbf{On the Object grid, PSG raises Reach from 11\% to 47\% at 20 to 30\,cm.} The released \piH\ and \piH\ + PSG on the Object grid: ten LIBERO-Object objects, each at its training place and at six new places; in percent of the episodes; \emph{all}: the object's episodes pooled over its places; \emph{cm}: how far the object was moved, with the episodes $N$ where fewer than ten; below, the moved places by distance, in episodes.}

\label{supp:C2}
\begin{tabular}{llw{r}{2.25em}w{r}{2.25em}w{r}{2.25em}w{r}{2.25em}w{r}{2.25em}w{r}{2.25em}w{r}{2.25em}w{r}{2.25em}w{r}{2.25em}w{r}{2.25em}w{r}{2.25em}w{r}{2.25em}w{r}{2.25em}w{r}{2.25em}w{r}{2.25em}w{r}{2.25em}}
\toprule
 & & \multicolumn{16}{c}{Our Object scenes, place of the requested object}\\
\cmidrule(lr){3-18}
Object & Method & \multicolumn{2}{c}{training} & \multicolumn{2}{c}{x2y3} & \multicolumn{2}{c}{x3y3} & \multicolumn{2}{c}{x4y3} & \multicolumn{2}{c}{x5y8} & \multicolumn{2}{c}{x6y6} & \multicolumn{2}{c}{x6y8} & \multicolumn{2}{c}{all}\\
 & & \ml{Reach} & \ml{Succ.} & \ml{Reach} & \ml{Succ.} & \ml{Reach} & \ml{Succ.} & \ml{Reach} & \ml{Succ.} & \ml{Reach} & \ml{Succ.} & \ml{Reach} & \ml{Succ.} & \ml{Reach} & \ml{Succ.} & \ml{Reach} & \ml{Succ.}\\
\midrule
alphabet soup & \piH & 100.0 & 100.0 & 90.0 & 60.0 & 100.0 & 70.0 & 100.0 & 30.0 & 0.0 & 0.0 & 0.0 & 0.0 & 0.0 & 0.0 & 55.7 & 37.1\\
\rowcolor[HTML]{EDEDED} & \ + PSG & 100.0 & 100.0 & 80.0 & 10.0 & 100.0 & 90.0 & 100.0 & 50.0 & 20.0 & 0.0 & 10.0 & 0.0 & 20.0 & 0.0 & 61.4 & 35.7\\
 & {\color{gray}\scriptsize\itshape cm} & \multicolumn{2}{c}{\color{gray}\scriptsize 0} & \multicolumn{2}{c}{\color{gray}\scriptsize 17} & \multicolumn{2}{c}{\color{gray}\scriptsize 11} & \multicolumn{2}{c}{\color{gray}\scriptsize 13} & \multicolumn{2}{c}{\color{gray}\scriptsize 42} & \multicolumn{2}{c}{\color{gray}\scriptsize 33} & \multicolumn{2}{c}{\color{gray}\scriptsize 47} & & \\
\addlinespace[2pt]
ketchup & \piH & 100.0 & 90.0 & 100.0 & 30.0 & 100.0 & 40.0 & 100.0 & 0.0 & 0.0 & 0.0 & 0.0 & 0.0 & 0.0 & 0.0 & 58.8 & 23.5\\
\rowcolor[HTML]{EDEDED} & \ + PSG & 100.0 & 100.0 & 100.0 & 70.0 & 100.0 & 30.0 & 100.0 & 10.0 & 12.5 & 0.0 & 0.0 & 0.0 & 0.0 & 0.0 & 60.3 & 30.9\\
 & {\color{gray}\scriptsize\itshape cm} & \multicolumn{2}{c}{\color{gray}\scriptsize 0} & \multicolumn{2}{c}{\color{gray}\scriptsize 17} & \multicolumn{2}{c}{\color{gray}\scriptsize 11} & \multicolumn{2}{c}{\color{gray}\scriptsize 13} & \multicolumn{2}{c}{\color{gray}\scriptsize 43; $N{=}8$} & \multicolumn{2}{c}{\color{gray}\scriptsize 33} & \multicolumn{2}{c}{\color{gray}\scriptsize 47} & & \\
\addlinespace[2pt]
butter & \piH & 100.0 & 100.0 & 90.0 & 90.0 & 100.0 & 100.0 & 100.0 & 90.0 & 0.0 & 0.0 & 0.0 & 0.0 & 0.0 & 0.0 & 55.7 & 54.3\\
\rowcolor[HTML]{EDEDED} & \ + PSG & 100.0 & 100.0 & 90.0 & 80.0 & 100.0 & 80.0 & 100.0 & 80.0 & 40.0 & 30.0 & 0.0 & 0.0 & 30.0 & 10.0 & 65.7 & 54.3\\
 & {\color{gray}\scriptsize\itshape cm} & \multicolumn{2}{c}{\color{gray}\scriptsize 0} & \multicolumn{2}{c}{\color{gray}\scriptsize 17} & \multicolumn{2}{c}{\color{gray}\scriptsize 12} & \multicolumn{2}{c}{\color{gray}\scriptsize 13} & \multicolumn{2}{c}{\color{gray}\scriptsize 42} & \multicolumn{2}{c}{\color{gray}\scriptsize 33} & \multicolumn{2}{c}{\color{gray}\scriptsize 47} & & \\
\addlinespace[2pt]
milk & \piH & 100.0 & 100.0 & 70.0 & 50.0 & 100.0 & 70.0 & 100.0 & 0.0 & 0.0 & 0.0 & 0.0 & 0.0 & 0.0 & 0.0 & 54.4 & 32.4\\
\rowcolor[HTML]{EDEDED} & \ + PSG & 100.0 & 100.0 & 90.0 & 60.0 & 100.0 & 40.0 & 100.0 & 0.0 & 0.0 & 0.0 & 0.0 & 0.0 & 0.0 & 0.0 & 57.4 & 29.4\\
 & {\color{gray}\scriptsize\itshape cm} & \multicolumn{2}{c}{\color{gray}\scriptsize 0} & \multicolumn{2}{c}{\color{gray}\scriptsize 17} & \multicolumn{2}{c}{\color{gray}\scriptsize 11} & \multicolumn{2}{c}{\color{gray}\scriptsize 13} & \multicolumn{2}{c}{\color{gray}\scriptsize 42; $N{=}8$} & \multicolumn{2}{c}{\color{gray}\scriptsize 33} & \multicolumn{2}{c}{\color{gray}\scriptsize 47} & & \\
\addlinespace[2pt]
chocolate pudding & \piH & 100.0 & 100.0 & 80.0 & 80.0 & 100.0 & 100.0 & 100.0 & 90.0 & 0.0 & 0.0 & 0.0 & 0.0 & 0.0 & 0.0 & 54.3 & 52.9\\
\rowcolor[HTML]{EDEDED} & \ + PSG & 100.0 & 100.0 & 100.0 & 100.0 & 100.0 & 90.0 & 100.0 & 80.0 & 0.0 & 0.0 & 20.0 & 10.0 & 10.0 & 10.0 & 61.4 & 55.7\\
 & {\color{gray}\scriptsize\itshape cm} & \multicolumn{2}{c}{\color{gray}\scriptsize 0} & \multicolumn{2}{c}{\color{gray}\scriptsize 17} & \multicolumn{2}{c}{\color{gray}\scriptsize 11} & \multicolumn{2}{c}{\color{gray}\scriptsize 13} & \multicolumn{2}{c}{\color{gray}\scriptsize 43} & \multicolumn{2}{c}{\color{gray}\scriptsize 33} & \multicolumn{2}{c}{\color{gray}\scriptsize 47} & & \\
\addlinespace[2pt]
cream cheese & \piH & 100.0 & 100.0 & 0.0 & 0.0 & 0.0 & 0.0 & 60.0 & 60.0 & 0.0 & 0.0 & 70.0 & 70.0 & 10.0 & 20.0 & 34.3 & 35.7\\
\rowcolor[HTML]{EDEDED} & \ + PSG & 100.0 & 100.0 & 0.0 & 0.0 & 0.0 & 10.0 & 100.0 & 100.0 & 40.0 & 30.0 & 100.0 & 80.0 & 100.0 & 90.0 & 62.9 & 58.6\\
 & {\color{gray}\scriptsize\itshape cm} & \multicolumn{2}{c}{\color{gray}\scriptsize 0} & \multicolumn{2}{c}{\color{gray}\scriptsize 39} & \multicolumn{2}{c}{\color{gray}\scriptsize 32} & \multicolumn{2}{c}{\color{gray}\scriptsize 27} & \multicolumn{2}{c}{\color{gray}\scriptsize 25} & \multicolumn{2}{c}{\color{gray}\scriptsize 11} & \multicolumn{2}{c}{\color{gray}\scriptsize 27} & & \\
\addlinespace[2pt]
salad dressing & \piH & 100.0 & 100.0 & 0.0 & 0.0 & 0.0 & 0.0 & 0.0 & 0.0 & 0.0 & 0.0 & 60.0 & 10.0 & 0.0 & 0.0 & 24.6 & 16.9\\
\rowcolor[HTML]{EDEDED} & \ + PSG & 100.0 & 100.0 & 0.0 & 0.0 & 0.0 & 0.0 & 10.0 & 10.0 & 80.0 & 20.0 & 100.0 & 40.0 & 60.0 & 0.0 & 47.7 & 24.6\\
 & {\color{gray}\scriptsize\itshape cm} & \multicolumn{2}{c}{\color{gray}\scriptsize 0} & \multicolumn{2}{c}{\color{gray}\scriptsize 39} & \multicolumn{2}{c}{\color{gray}\scriptsize 32} & \multicolumn{2}{c}{\color{gray}\scriptsize 27} & \multicolumn{2}{c}{\color{gray}\scriptsize 25; $N{=}5$} & \multicolumn{2}{c}{\color{gray}\scriptsize 11} & \multicolumn{2}{c}{\color{gray}\scriptsize 27} & & \\
\addlinespace[2pt]
BBQ sauce & \piH & 100.0 & 100.0 & 0.0 & 0.0 & 0.0 & 0.0 & 0.0 & 0.0 & 0.0 & 0.0 & 50.0 & 10.0 & 0.0 & 0.0 & 21.4 & 15.7\\
\rowcolor[HTML]{EDEDED} & \ + PSG & 100.0 & 100.0 & 0.0 & 0.0 & 0.0 & 0.0 & 0.0 & 0.0 & 10.0 & 0.0 & 100.0 & 50.0 & 0.0 & 0.0 & 30.0 & 21.4\\
 & {\color{gray}\scriptsize\itshape cm} & \multicolumn{2}{c}{\color{gray}\scriptsize 0} & \multicolumn{2}{c}{\color{gray}\scriptsize 38} & \multicolumn{2}{c}{\color{gray}\scriptsize 31} & \multicolumn{2}{c}{\color{gray}\scriptsize 26} & \multicolumn{2}{c}{\color{gray}\scriptsize 25} & \multicolumn{2}{c}{\color{gray}\scriptsize 12} & \multicolumn{2}{c}{\color{gray}\scriptsize 27} & & \\
\addlinespace[2pt]
tomato sauce & \piH & 100.0 & 100.0 & 0.0 & 0.0 & 0.0 & 0.0 & 0.0 & 0.0 & 37.5 & 0.0 & 100.0 & 50.0 & 50.0 & 0.0 & 41.2 & 22.1\\
\rowcolor[HTML]{EDEDED} & \ + PSG & 100.0 & 100.0 & 0.0 & 0.0 & 0.0 & 0.0 & 0.0 & 0.0 & 100.0 & 12.5 & 100.0 & 20.0 & 100.0 & 0.0 & 55.9 & 19.1\\
 & {\color{gray}\scriptsize\itshape cm} & \multicolumn{2}{c}{\color{gray}\scriptsize 0} & \multicolumn{2}{c}{\color{gray}\scriptsize 39} & \multicolumn{2}{c}{\color{gray}\scriptsize 32} & \multicolumn{2}{c}{\color{gray}\scriptsize 27} & \multicolumn{2}{c}{\color{gray}\scriptsize 25; $N{=}8$} & \multicolumn{2}{c}{\color{gray}\scriptsize 11} & \multicolumn{2}{c}{\color{gray}\scriptsize 27} & & \\
\addlinespace[2pt]
orange juice & \piH & 100.0 & 100.0 & 0.0 & 0.0 & 0.0 & 0.0 & 10.0 & 0.0 & 0.0 & 0.0 & 80.0 & 10.0 & 0.0 & 0.0 & 27.1 & 15.7\\
\rowcolor[HTML]{EDEDED} & \ + PSG & 100.0 & 100.0 & 0.0 & 0.0 & 10.0 & 0.0 & 50.0 & 10.0 & 70.0 & 20.0 & 100.0 & 10.0 & 10.0 & 0.0 & 48.6 & 20.0\\
 & {\color{gray}\scriptsize\itshape cm} & \multicolumn{2}{c}{\color{gray}\scriptsize 0} & \multicolumn{2}{c}{\color{gray}\scriptsize 39} & \multicolumn{2}{c}{\color{gray}\scriptsize 32} & \multicolumn{2}{c}{\color{gray}\scriptsize 27} & \multicolumn{2}{c}{\color{gray}\scriptsize 25} & \multicolumn{2}{c}{\color{gray}\scriptsize 11} & \multicolumn{2}{c}{\color{gray}\scriptsize 27} & & \\
\midrule all ten & \piH & 100.0 & 99.0 & 43.0 & 31.0 & 50.0 & 38.0 & 57.0 & 27.0 & 3.4 & 0.0 & 36.0 & 15.0 & 6.0 & 2.0 & 42.8 & 30.8\\
\rowcolor[HTML]{EDEDED} & \ + PSG & 100.0 & 100.0 & 46.0 & 32.0 & 51.0 & 34.0 & 66.0 & 34.0 & 34.8 & 11.2 & 53.0 & 21.0 & 33.0 & 11.0 & 55.2 & 35.1\\
\bottomrule
\end{tabular}

\par\medskip
\begin{tabular}{lrrrrr}
\toprule
 & & \multicolumn{2}{c}{\piH} & \multicolumn{2}{c}{\piH\ + PSG} \\
\cmidrule(lr){3-4}\cmidrule(lr){5-6}
Object moved & \ml{Ep.} & \ml{reaches} & \ml{succ.} & \ml{reaches} & \ml{succ.} \\
\midrule
under 20\,cm & 200 & 179 & 105 & 196 & 107 \\
20 to 30\,cm & 143 & 16 & 8 & 67 & 28 \\
over 30\,cm & 246 & 0 & 0 & 17 & 7 \\
\bottomrule
\end{tabular}

\end{table*}
{\small C2: not in place yet.}}\par}
\begin{table*}[tp]
\centering
\setlength\heavyrulewidth{0.25ex}
\captionsetup{position=top,skip=4pt}
\caption{\textbf{LIBERO-Goal with the target object moved, per scene.} Reach (\textit{R}) and success (\textit{S}) for each of the three Goal objects at its training place and at six new places, by distance moved, 10 episodes per scene; bold: the groups of the main paper's Tab.~2. Gray: requires training ($\dagger$).}

\label{supp:D1e}
\footnotesize
\setlength{\tabcolsep}{2.4pt}
\begin{tabular}{lr@{\hspace{1.5pt}}r@{\hspace{6pt}}r@{\hspace{1.5pt}}r@{\hspace{6pt}}r@{\hspace{1.5pt}}r@{\hspace{6pt}}r@{\hspace{1.5pt}}r@{\hspace{6pt}}r@{\hspace{1.5pt}}r@{\hspace{6pt}}>{\color{black!55}}r@{\hspace{1.5pt}}>{\color{black!55}}r@{\hspace{6pt}}>{\color{black!55}}r@{\hspace{1.5pt}}>{\color{black!55}}r@{\hspace{6pt}}r@{\hspace{1.5pt}}r@{\hspace{6pt}}r@{\hspace{1.5pt}}r@{\hspace{6pt}}>{\columncolor[HTML]{EDEDED}[3pt][0.75pt]}r@{\hspace{1.5pt}}>{\columncolor[HTML]{EDEDED}[0.75pt][3pt]}r}
\toprule
Task & \multicolumn{2}{c}{OFT} & \multicolumn{2}{c}{SmolVLA} & \multicolumn{2}{c}{TurboVLA} & \multicolumn{2}{c}{\piO-FAST} & \multicolumn{2}{c}{\piO} & \multicolumn{2}{c}{RSS$^\dagger$} & \multicolumn{2}{c}{APT$^\dagger$} & \multicolumn{2}{c}{\piH} & \multicolumn{2}{c}{+\,CAG} & \multicolumn{2}{c}{+\,PSG}\\
\cmidrule(lr){2-3}\cmidrule(lr){4-5}\cmidrule(lr){6-7}\cmidrule(lr){8-9}\cmidrule(lr){10-11}\cmidrule(lr){12-13}\cmidrule(lr){14-15}\cmidrule(lr){16-17}\cmidrule(lr){18-19}\cmidrule(lr){20-21}
 & {\scriptsize\itshape R} & {\scriptsize\itshape S} & {\scriptsize\itshape R} & {\scriptsize\itshape S} & {\scriptsize\itshape R} & {\scriptsize\itshape S} & {\scriptsize\itshape R} & {\scriptsize\itshape S} & {\scriptsize\itshape R} & {\scriptsize\itshape S} & {\scriptsize\itshape R} & {\scriptsize\itshape S} & {\scriptsize\itshape R} & {\scriptsize\itshape S} & {\scriptsize\itshape R} & {\scriptsize\itshape S} & {\scriptsize\itshape R} & {\scriptsize\itshape S} & {\scriptsize\itshape R} & {\scriptsize\itshape S}\\
\midrule
\multicolumn{21}{l}{\textit{LIBERO-Goal, target object moved}}\\
\parbox[t]{3.4cm}{\raggedright bowl, training place} & 100 & 90 & 100 & 100 & 100 & 100 & 100 & 90 & 100 & 100 & 100 & 100 & 100 & 100 & 100 & 100 & 100 & 100 & 100 & 100\\
\parbox[t]{3.4cm}{\raggedright bowl, x1y9, 47.5 cm} & 0 & 0 & 0 & 0 & 0 & 0 & 0 & 0 & 0 & 0 & 70 & 20 & 0 & 0 & 0 & 0 & 0 & 0 & 0 & 20\\
\parbox[t]{3.4cm}{\raggedright bowl, x2y6, 22.5 cm} & 0 & 0 & 0 & 0 & 0 & 0 & 40 & 0 & 0 & 0 & 100 & 80 & 50 & 20 & 20 & 50 & 100 & 50 & 100 & 90\\
\parbox[t]{3.4cm}{\raggedright bowl, x2y9, 42.8 cm} & 0 & 0 & 0 & 0 & 0 & 0 & 0 & 0 & 0 & 0 & 100 & 100 & 30 & 0 & 0 & 0 & 20 & 80 & 30 & 80\\
\parbox[t]{3.4cm}{\raggedright bowl, x5y8, 26.5 cm} & 0 & 0 & 0 & 0 & 0 & 0 & 70 & 0 & 0 & 0 & 100 & 100 & 70 & 0 & 100 & 90 & 100 & 90 & 100 & 80\\
\parbox[t]{3.4cm}{\raggedright bowl, x6y7, 23.7 cm} & 10 & 0 & 0 & 0 & 20 & 0 & 90 & 0 & 0 & 0 & 100 & 90 & 70 & 0 & 100 & 90 & 100 & 60 & 100 & 50\\
\parbox[t]{3.4cm}{\raggedright bowl, x6y8, 31.1 cm} & 0 & 0 & 0 & 0 & 0 & 0 & 30 & 0 & 0 & 0 & 100 & 70 & 10 & 0 & 60 & 60 & 60 & 50 & 70 & 30\\
\parbox[t]{3.4cm}{\raggedright cream cheese, training place} & 100 & 100 & 100 & 50 & 100 & 100 & 100 & 90 & 100 & 100 & 100 & 100 & 100 & 100 & 100 & 100 & 100 & 100 & 100 & 100\\
\parbox[t]{3.4cm}{\raggedright cream cheese, x1y9, 42.0 cm} & 0 & 0 & 0 & 0 & 0 & 0 & 0 & 0 & 0 & 0 & 0 & 0 & 0 & 0 & 0 & 0 & 0 & 0 & 0 & 0\\
\parbox[t]{3.4cm}{\raggedright cream cheese, x2y6, 26.9 cm} & 0 & 0 & 0 & 0 & 0 & 0 & 0 & 0 & 0 & 0 & 100 & 90 & 80 & 20 & 0 & 0 & 20 & 60 & 100 & 80\\
\parbox[t]{3.4cm}{\raggedright cream cheese, x2y9, 33.9 cm} & 0 & 0 & 0 & 0 & 0 & 0 & 0 & 0 & 0 & 0 & 0 & 0 & 30 & 10 & 0 & 0 & 0 & 0 & 10 & 0\\
\parbox[t]{3.4cm}{\raggedright cream cheese, x5y8, 12.9 cm} & 0 & 0 & 10 & 0 & 0 & 0 & 100 & 50 & 10 & 30 & 100 & 100 & 100 & 70 & 100 & 90 & 100 & 100 & 100 & 100\\
\parbox[t]{3.4cm}{\raggedright cream cheese, x6y7, 14.5 cm} & 20 & 0 & 0 & 0 & 10 & 0 & 100 & 0 & 70 & 0 & 100 & 10 & 100 & 70 & 100 & 0 & 100 & 10 & 100 & 70\\
\parbox[t]{3.4cm}{\raggedright cream cheese, x6y8, 18.8 cm} & 0 & 0 & 0 & 0 & 0 & 0 & 100 & 20 & 0 & 0 & 100 & 30 & 100 & 40 & 100 & 0 & 100 & 0 & 100 & 70\\
\parbox[t]{3.4cm}{\raggedright wine bottle, training place} & 100 & 100 & 100 & 80 & 100 & 100 & 100 & 100 & 100 & 100 & 100 & 100 & 100 & 100 & 100 & 100 & 100 & 100 & 100 & 100\\
\parbox[t]{3.4cm}{\raggedright wine bottle, x1y9, 45.3 cm} & 0 & 0 & 0 & 0 & 0 & 0 & 0 & 0 & 0 & 0 & 0 & 0 & 0 & 0 & 0 & 0 & 0 & 0 & 0 & 0\\
\parbox[t]{3.4cm}{\raggedright wine bottle, x2y6, 15.1 cm} & 0 & 0 & 0 & 0 & 0 & 0 & 20 & 0 & 90 & 10 & 100 & 70 & 100 & 50 & 100 & 100 & 100 & 70 & 100 & 100\\
\parbox[t]{3.4cm}{\raggedright wine bottle, x2y9, 41.5 cm} & 0 & 0 & 0 & 0 & 0 & 0 & 0 & 0 & 0 & 0 & 10 & 0 & 0 & 0 & 0 & 0 & 0 & 0 & 100 & 0\\
\parbox[t]{3.4cm}{\raggedright wine bottle, x5y8, 35.3 cm} & 0 & 0 & 0 & 0 & 0 & 0 & 0 & 0 & 0 & 0 & 40 & 10 & 50 & 30 & 70 & 0 & 30 & 0 & 100 & 0\\
\parbox[t]{3.4cm}{\raggedright wine bottle, x6y7, 35.0 cm} & 0 & 0 & 0 & 0 & 0 & 0 & 0 & 0 & 0 & 0 & 60 & 0 & 0 & 0 & 40 & 0 & 10 & 0 & 90 & 0\\
\parbox[t]{3.4cm}{\raggedright wine bottle, x6y8, 41.5 cm} & 0 & 0 & 0 & 0 & 0 & 0 & 0 & 0 & 0 & 0 & 80 & 0 & 0 & 0 & 70 & 0 & 100 & 0 & 100 & 0\\
\parbox[t]{3.4cm}{\raggedright \textbf{$<$ 20 cm}} & 5.0 & 0.0 & 2.5 & 0.0 & 2.5 & 0.0 & 80.0 & 17.5 & 42.5 & 10.0 & 100.0 & 52.5 & 100.0 & 57.5 & 100.0 & 47.5 & 100.0 & 45.0 & 100.0 & 85.0\\
\parbox[t]{3.4cm}{\raggedright \textbf{20-30 cm}} & 2.5 & 0.0 & 0.0 & 0.0 & 5.0 & 0.0 & 50.0 & 0.0 & 0.0 & 0.0 & 100.0 & 90.0 & 67.5 & 10.0 & 55.0 & 57.5 & 80.0 & 65.0 & 100.0 & 75.0\\
\parbox[t]{3.4cm}{\raggedright \textbf{$>$ 30 cm}} & 0 & 0 & 0 & 0 & 0 & 0 & 3 & 0 & 0 & 0 & 46 & 20 & 12 & 4 & 24 & 6 & 22 & 13 & 50 & 13\\
\bottomrule
\end{tabular}

\end{table*}
{\small D1e: not in place yet.}}\par}
\begin{table*}[tp]
\centering
\setlength\heavyrulewidth{0.25ex}
\captionsetup{position=top,skip=4pt}
\caption{\textbf{LIBERO-CF per task.} Reach (\textit{R}) and success (\textit{S}) of the rows of the main paper's Tab.~3, 50 episodes per task; \textbf{all tasks}: its column. Gray: requires training ($\dagger$).}

\label{supp:D1f}
\footnotesize
\setlength{\tabcolsep}{2.4pt}
\begin{tabular}{lr@{\hspace{1.5pt}}r@{\hspace{6pt}}>{\color{black!55}}r@{\hspace{1.5pt}}>{\color{black!55}}r@{\hspace{6pt}}r@{\hspace{1.5pt}}r@{\hspace{6pt}}>{\columncolor[HTML]{EDEDED}[3pt][0.75pt]}r@{\hspace{1.5pt}}>{\columncolor[HTML]{EDEDED}[0.75pt][3pt]}r}
\toprule
Task & \multicolumn{2}{c}{\piH} & \multicolumn{2}{c}{+\,CAG-VA$^\dagger$} & \multicolumn{2}{c}{+\,CAG} & \multicolumn{2}{c}{+\,PSG}\\
\cmidrule(lr){2-3}\cmidrule(lr){4-5}\cmidrule(lr){6-7}\cmidrule(lr){8-9}
 & {\scriptsize\itshape R} & {\scriptsize\itshape S} & {\scriptsize\itshape R} & {\scriptsize\itshape S} & {\scriptsize\itshape R} & {\scriptsize\itshape S} & {\scriptsize\itshape R} & {\scriptsize\itshape S}\\
\midrule
\multicolumn{9}{l}{\textit{LIBERO-CF Spatial}}\\
\parbox[t]{7.2cm}{\raggedright 01 pick up the cookie box and place it on the plate} & 90 & 60 & 94 & 46 & 98 & 56 & 78 & 50\\
\parbox[t]{7.2cm}{\raggedright 02 pick up the ramekin and place it on the plate} & 0 & 0 & 0 & 0 & 0 & 0 & 0 & 0\\
\parbox[t]{7.2cm}{\raggedright 03 pick up the cookie box and place it on the plate} & 24 & 22 & 0 & 0 & 40 & 36 & 46 & 36\\
\parbox[t]{7.2cm}{\raggedright 04 pick up the cookie box and place it on the plate} & 14 & 12 & 40 & 36 & 42 & 32 & 40 & 16\\
\parbox[t]{7.2cm}{\raggedright 05 pick up the ramekin and place it on the plate} & 0 & 0 & 2 & 2 & 0 & 0 & 0 & 0\\
\parbox[t]{7.2cm}{\raggedright 06 pick up the ramekin and place it on the plate} & 2 & 2 & 8 & 8 & 0 & 0 & 0 & 0\\
\parbox[t]{7.2cm}{\raggedright 07 pick up the black bowl on the cookie box and place it on the plate} & 96 & 92 & 100 & 98 & 100 & 100 & 100 & 96\\
\parbox[t]{7.2cm}{\raggedright 08 pick up the cookie box and place it on the plate} & 4 & 2 & 32 & 22 & 46 & 32 & 48 & 38\\
\parbox[t]{7.2cm}{\raggedright 09 pick up the cookie box and place it on the plate} & 100 & 62 & 100 & 40 & 100 & 72 & 92 & 62\\
\parbox[t]{7.2cm}{\raggedright 10 pick up the black bowl next to the ramekin and place it on the plate} & 32 & 26 & 100 & 92 & 92 & 86 & 30 & 30\\
\parbox[t]{7.2cm}{\raggedright 11 pick up the ramekin and place it on the plate} & 0 & 0 & 0 & 0 & 0 & 0 & 0 & 2\\
\parbox[t]{7.2cm}{\raggedright 12 pick up the black bowl next to the plate and place it on the plate} & 4 & 6 & 8 & 8 & 8 & 8 & 4 & 4\\
\parbox[t]{7.2cm}{\raggedright 13 pick up the black bowl on the wooden cabinet and place it on the plate} & 82 & 80 & 98 & 98 & 98 & 92 & 94 & 88\\
\parbox[t]{7.2cm}{\raggedright 14 pick up the black bowl on the wooden cabinet and place it on the plate} & 6 & 6 & 44 & 38 & 4 & 4 & 50 & 42\\
\parbox[t]{7.2cm}{\raggedright 15 pick up the ramekin and place it on the plate} & 0 & 0 & 0 & 0 & 0 & 0 & 0 & 0\\
\parbox[t]{7.2cm}{\raggedright \textbf{all tasks}} & 30.3 & 24.7 & 41.7 & 32.5 & 41.9 & 34.5 & 38.8 & 30.9\\
\bottomrule
\end{tabular}

\end{table*}
{\small D1f: not in place yet.}}\par}
\twocolumn[{\suppblock{D1last}{\textwidth}{\vspace{10pt}}}]
\section{Limitations}
\label{supp:limits}
\noindent\textbf{Scope.}
The pathway cuts are applied to \piH\ and \piO; PSG is evaluated on \piH, with one control run on \piO\ (\cref{supp:C6}).
All scenes come from LIBERO and its perturbed versions and run in simulation with one robot arm, and the main paper's tables use one noise seed (\cref{supp:D2}).

\noindent\textbf{Pathway cuts.}
The two cuts isolate two parts of the instruction's influence with large and opposite effects under Position.
They do not separate grounding from the shortcut completely: instruction information that has passed into the image tokens, and the image itself, may also draw the hand to a training place.
Nor is the direct pathway independent of the image: in the bidirectional prefix, the instruction tokens that the action tokens read have themselves read the image.

\noindent\textbf{PSG.}
PSG amplifies grounding that the policy already has: where the name's attention does not single out the object, as for the BBQ sauce, it adds no reach (\cref{supp:C3}), and beyond 30\,cm on the Object grid it adds few reaches (\cref{supp:C2}).
PSG's weights were selected on eleven development tasks whose start states 6 to 15 are among the evaluated episodes (\cref{supp:protocol-text}).
[E0: text to come.]}\par}
\clearpage

\end{document}